\documentclass[11pt, a4paper, copyright]{google}

\usepackage[numbers,compress]{natbib}
\usepackage{hyperref}       
\usepackage{url}            
\usepackage{booktabs}       
\usepackage{amsfonts}       
\usepackage{nicefrac}       
\usepackage{microtype}      
\usepackage[dvipsnames]{xcolor}         
\usepackage{lipsum} 
\usepackage{amsmath}
\usepackage{bbm}
\usepackage{amssymb} 
\usepackage{multirow}
\usepackage{relsize}
\usepackage{array}
\usepackage{colortbl}
\usepackage{pifont}
\usepackage{hhline}
\usepackage{boldline}
\usepackage{float}
\usepackage{pifont}
\usepackage{subcaption}
\usepackage[figuresright]{rotating}
\usepackage{balance}
\usepackage{blindtext}
\usepackage{paralist}
\usepackage{arydshln} 
\usepackage{enumitem}
\usepackage{wrapfig}
\usepackage{colortbl}
\usepackage[most]{tcolorbox}
\tcbuselibrary{listings,breakable}
\usepackage{graphicx}
\usepackage{caption}
\usepackage[ruled,vlined]{algorithm2e}

\usepackage{xspace}
\usepackage{setspace}
\newcommand{\ourspace}{{\fontfamily{lmtt}\selectfont \textbf{GenCanvas}}\xspace}
\newcommand{\ourmethod}{{\fontfamily{lmtt}\selectfont \textbf{GenRouter}}\xspace}

\definecolor{mygrey}{gray}{0.4}

\definecolor{lightpurple}{RGB}{242, 238, 255} 
\definecolor{lightyellow}{RGB}{255, 252, 220}

\newcommand{\qwenlogo}{\raisebox{-0.15em}{\includegraphics[height=1em]{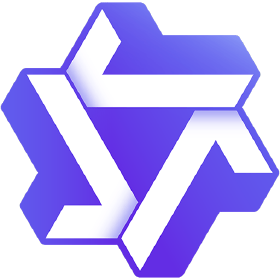}}}
\newcommand{\zlogo}{\raisebox{-0.15em}{\includegraphics[height=1em]{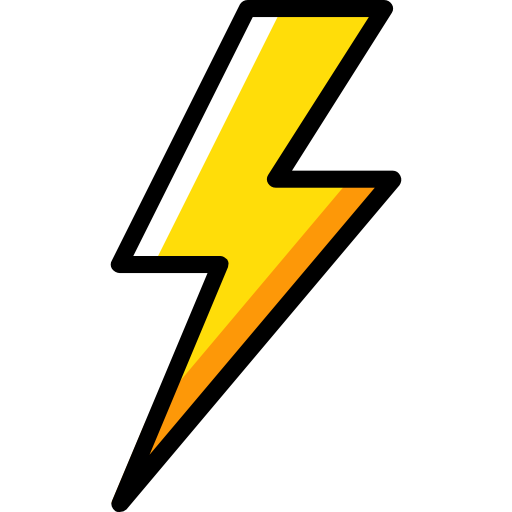}}}

\hypersetup{
    colorlinks=true,
    linkcolor=magenta,
    citecolor=cyan,
    filecolor=magenta,      
    urlcolor=magenta,
    }

\newcommand{\obsbox}[1]{%
    \begin{tcolorbox}[colframe=black!70, colback=cyan!2, boxrule=1pt, arc=2mm,   top=3pt, bottom=3pt, left=3pt, right=3pt,
  boxsep=1pt]
        #1
    \end{tcolorbox}
}

\definecolor{linkred}{HTML}{a33a32}
\usepackage{nicefrac}       
\usepackage[most]{tcolorbox}

\newcommand{\answerTODO}[1][]{\textcolor{red}{\bf [TODO]}}

\uselogo{}
\correspondingauthor{}
\renewcommand{\copyrightext}{{\footerfont\itshape $^*$Equal Contribution. $^{\dagger}$Corresponding Author.~~~~~~~~~~~~~~~~~~~~~~~~~~~~~~~~~~~~~~~~~~~~~~~~~~~~~~~~~~~~~~~~~~~~~~~~~~~Primary Contact: haroldchen328@gmail.com}}

\makeatletter
\renewcommand{\absfont}{\normalfont\linespread{1.2}\fontsize{11}{12}\selectfont}
\renewcommand{\titlefont}{\color{black}\normalfont\bfseries\fontsize{18.5}{23}\selectfont}
\makeatother

\title{\ourmethod: Unified Workflow Routing for Agentic Image Generation}

\author{
Harold Haodong Chen$^{1,2,*}$,
Zhiyu Hou$^{1,3,*}$,
Wen-Jie Shu$^{4}$,
Weilin Ruan$^{5}$,
Yingjie Xu$^{1}$, ~~~~~
Litao Guo$^{1}$,
Ying-Cong Chen$^{1,2,\dagger}$ \\
$^{1}$HKUST(GZ), $^{2}$HKUST, $^{3}$SUSTech, $^{4}$ZODA, $^{5}$CUHK\\
\vspace{0.4em}
{\fontsize{11pt}{11pt} \selectfont \raisebox{-0.06em}{\includegraphics[height=1em]{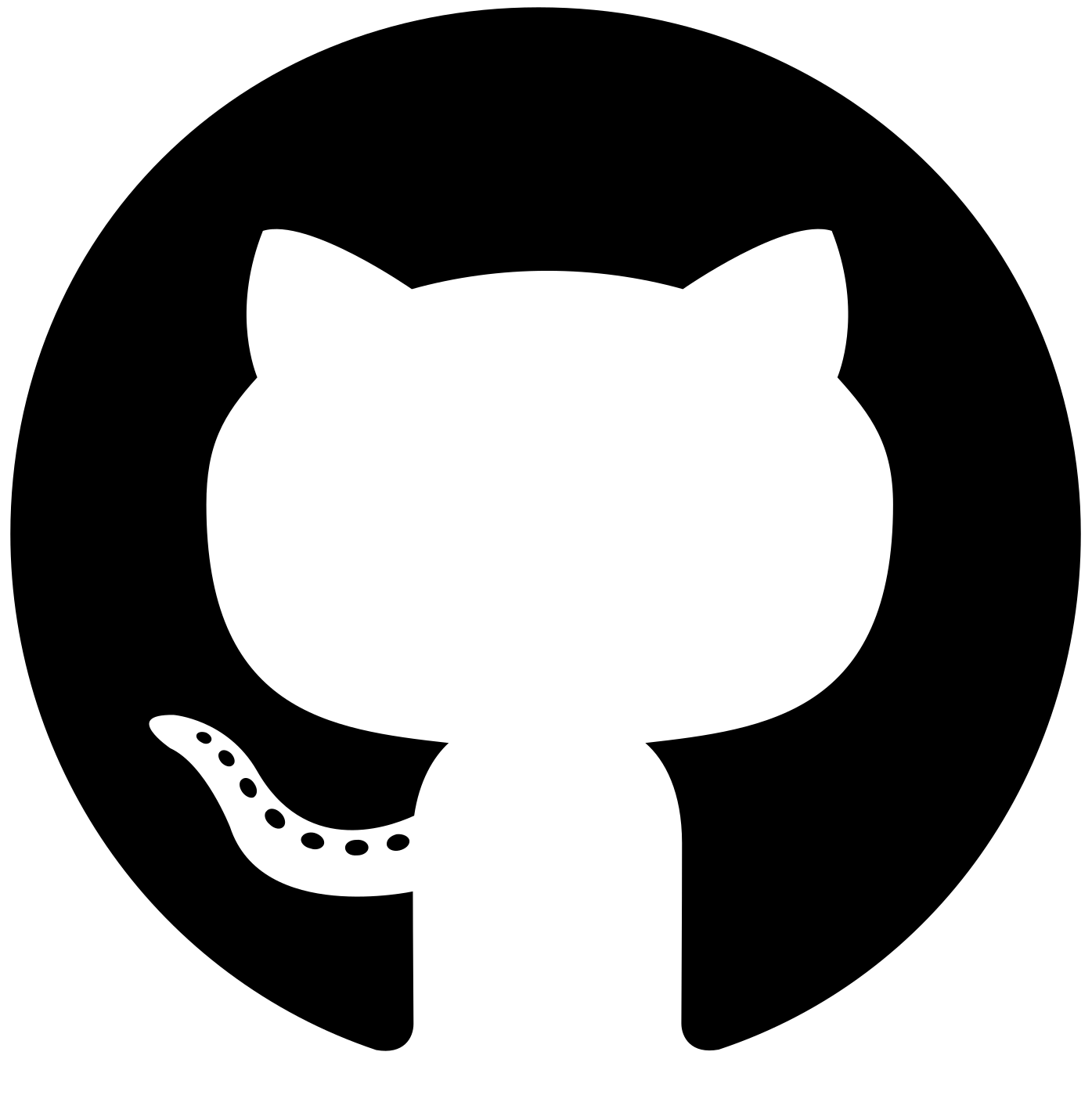}} \href{https://github.com/EnVision-Research/GenRouter}{https://github.com/EnVision-Research/GenRouter}}
}

\usepackage{amsmath,amsfonts,bm}

\usepackage{cleveref}
\usepackage{wrapfig}
\usepackage{caption}

\usepackage[most]{tcolorbox}
\usepackage[table]{xcolor}
\usepackage{listings}
\usepackage[T1]{fontenc}
\usepackage{subcaption}
\usepackage{makecell}
\usepackage{multirow}
\usepackage{twemojis}

\def\eqref#1{equation~\ref{#1}}

\def\1{\bm{1}}

\DeclareMathAlphabet{\mathsfit}{\encodingdefault}{\sfdefault}{m}{sl}
\SetMathAlphabet{\mathsfit}{bold}{\encodingdefault}{\sfdefault}{bx}{n}

\definecolor{lightpink}{HTML}{f7ded8}
\definecolor{lightyellow}{HTML}{f7eede}
\definecolor{lightgreen}{HTML}{DDEBDD}

\definecolor{pink}{HTML}{d85a4a}
\definecolor{yellow}{HTML}{c6a15b}
\definecolor{green}{HTML}{789c80}
\definecolor{red}{HTML}{9A3F3A}

\newtcolorbox{prompt}[3][]{
    colback=#2,
    colbacktitle=#3,
    colframe=#3,
    coltitle=white,
    fontupper=\small\ttfamily,
    fonttitle=\small\bfseries,
    boxsep=2pt,
    left=0pt,
    right=0pt,
    top=0pt,
    bottom=0pt,
    boxrule=0.8pt,
    enhanced,
    breakable,
    #1,
}

\newtcblisting{promptlisting}[2][]{
    enhanced,
    breakable,
    colback=lightgreen!50,
    colbacktitle=green,
    colframe=green,
    coltitle=white,
    fonttitle=\small\bfseries,
    boxsep=2pt,
    left=4pt,
    right=4pt,
    top=2pt,
    bottom=2pt,
    boxrule=0.6pt,
    listing only,
    listing options={
        basicstyle=\scriptsize\ttfamily,
        breaklines=true,
        breakatwhitespace=false,
        columns=fullflexible,
        keepspaces=true,
        showstringspaces=false,
        breakindent=0pt,
    },
    title={#2},
    #1,
}

\newtcblisting{examplelisting}[2][]{
    enhanced,
    breakable,
    colback=lightyellow!50,
    colbacktitle=yellow!85,
    colframe=yellow!85,
    coltitle=white,
    fonttitle=\small\bfseries,
    boxsep=2pt,
    left=4pt,
    right=4pt,
    top=2pt,
    bottom=2pt,
    boxrule=0.6pt,
    listing only,
    listing options={
        basicstyle=\scriptsize\ttfamily,
        breaklines=true,
        breakatwhitespace=false,
        columns=fullflexible,
        keepspaces=true,
        showstringspaces=false,
        breakindent=0pt,
    },
    title={#2},
    #1,
}

\makeatletter

\renewcommand{\maketitle}{\bgroup\setlength{\parindent}{0pt}
	\begin{adjustwidth}{0pt}{24pt}
		\begin{flushleft}
			{
				{\raggedright \titlefont \@title\par}%
				\vskip11pt
				{\raggedright \@author\par}
				\vskip20pt%
			}%
		\end{flushleft}
	\end{adjustwidth}
	\egroup
	\thispagestyle{firststyle} 
	{%
		{\abscontent}
	}%
}%

\renewcommand{\abscontent}{
    \begingroup
    \centering
    \vspace{-1.6em}
    \includegraphics[width=\linewidth]{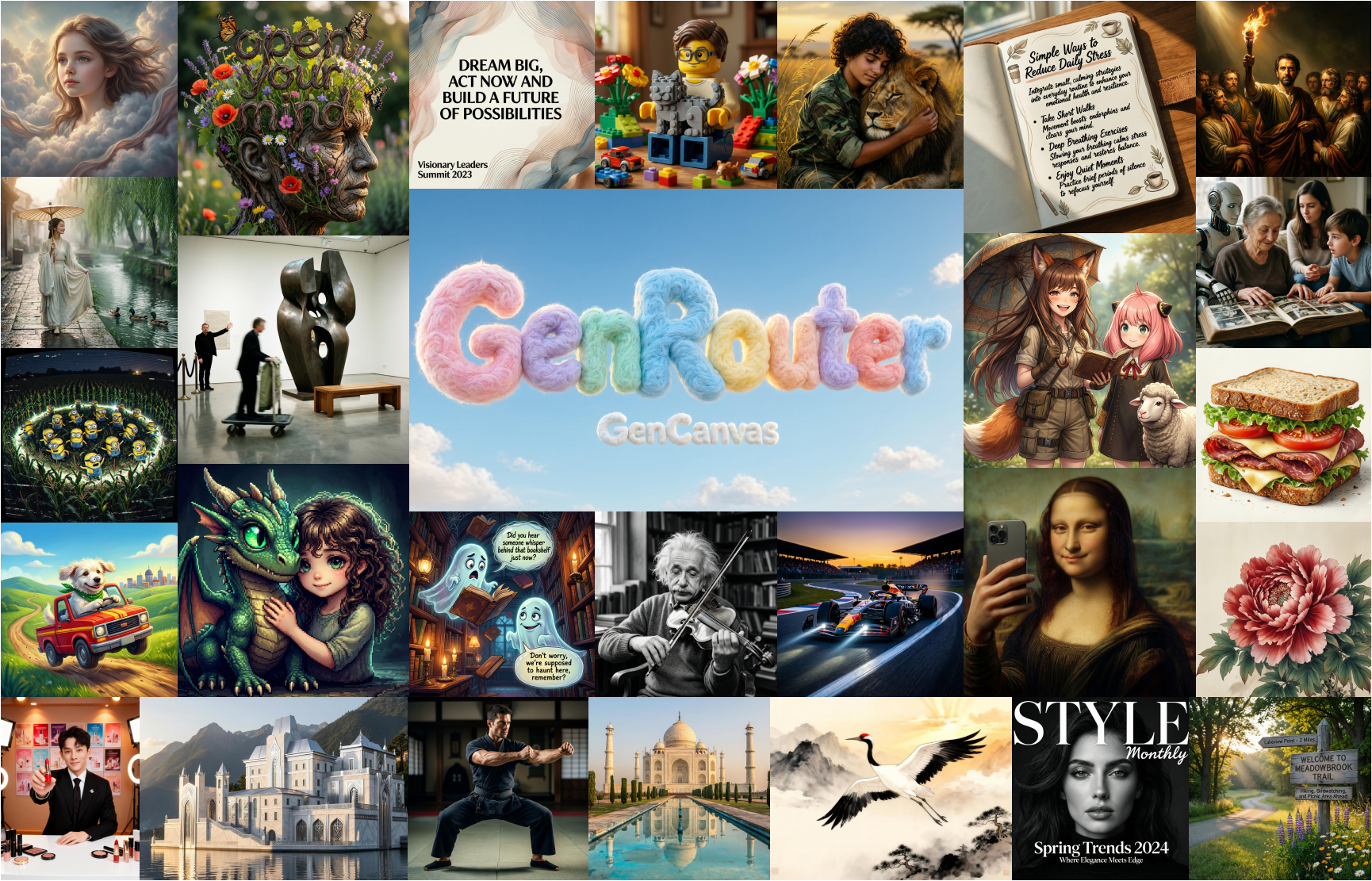}\par
    \vspace{-0.6em}
    \captionof{figure}{Generated images using our proposed \ourmethod within \ourspace.}
    \label{fig:teaser}\par
    \vspace{0.6em}
    \endgroup
    
    \noindent
    {\absfont \theabstract \par}
    
    \@ifundefined{@keywords}{}{
        \vskip1em \noindent \keywordsfont  Keywords: \@keywords}
}
\makeatother

\begin{document}

\begin{abstract}
\textbf{Abstract.}\quad The rapid evolution of text-to-image (T2I) generation models has effectively solved the foundational challenge of raw pixel synthesis, shifting the community's focus toward fulfilling increasingly intricate user requests. While recent agentic image generation workflows enhance static inference with advanced capabilities like external knowledge retrieval and iterative reasoning, they mostly operate in isolated silos with fixed ``one-size-fits-all" topologies. This inevitably leads to severe compute-mismatch, where simple queries are forced through computationally heavy pipelines. To bridge this gap, we present \ourmethod, the first unified workflow routing framework for agentic image generation. We first formulate \ourspace, standardizing diverse agentic pipelines into a universal set of foundational primitives and executable templates. Operating over this unified space, \ourmethod adaptively routes heterogeneous prompts to their optimal workflows via (\textbf{\textit{i}})~\textbf{demand profiling}, (\textbf{\textit{ii}})~\textbf{experience matching}, and (\textbf{\textit{iii}})~\textbf{Pareto filtering}. Extensive experiments across diverse benchmarks demonstrate that \ourmethod achieves superior visual alignment while reducing execution costs by over $95\%$ and latency by $65\%$ compared to heavyweight static pipelines. Furthermore, the system continuously self-evolves via accumulated experience, enabling robust zero-shot generalization that boosts performance and halves computational overhead.

\end{abstract} 

\maketitle

\section{Introduction}

\begin{wrapfigure}{r}{0.5\textwidth}
\vspace{-1.6em}
 \centering
 \includegraphics[width=\linewidth]{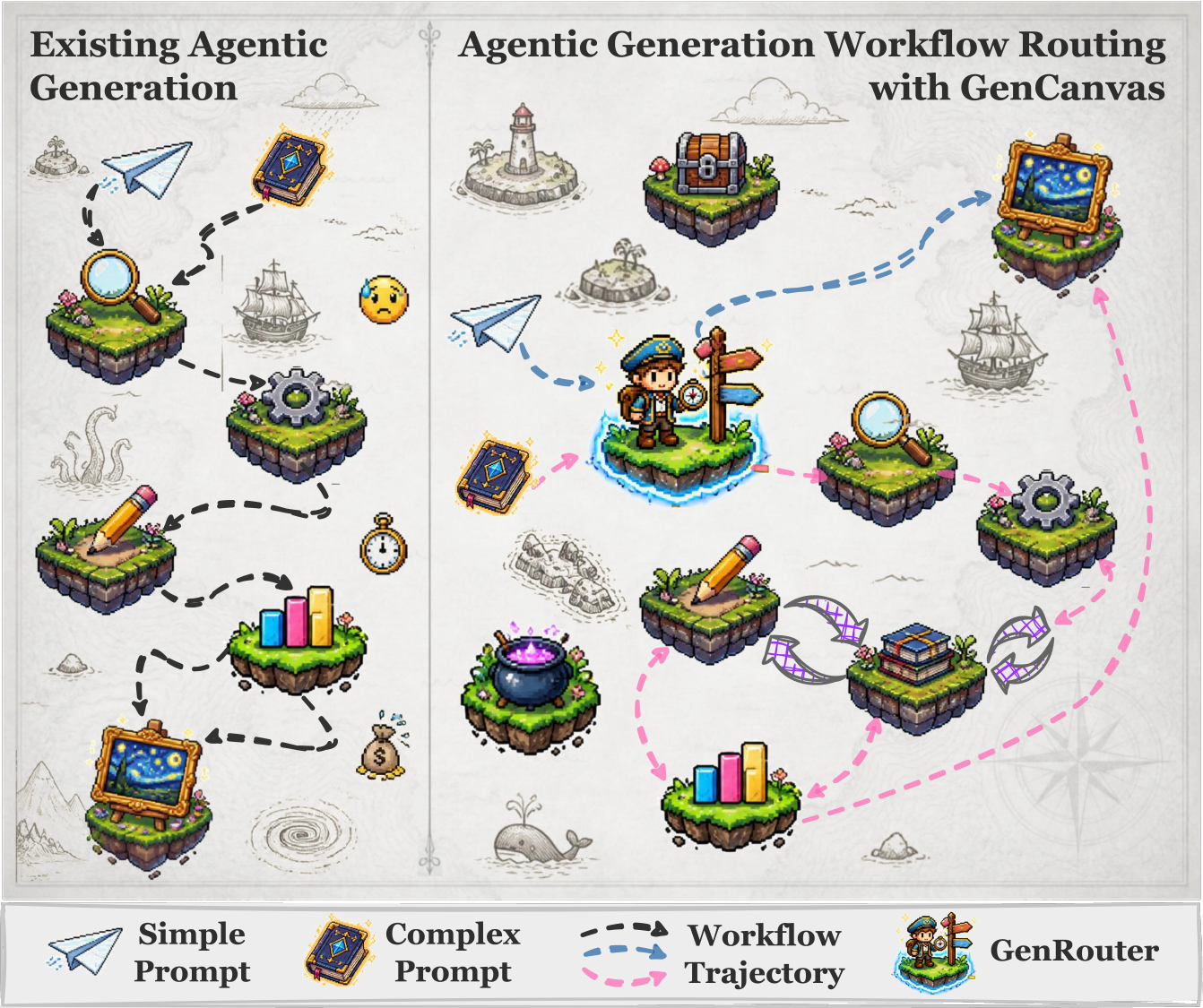}
  \vspace{-2em}
  \caption{(\textbf{\textit{Left}}) Existing systems enforce a fixed execution workflow. (\textbf{\textit{Right}}) \ourmethod dynamically routes prompts to their optimal workflow trajectories over the unified \ourspace.}
  \vspace{-2em}
  \label{fig:figure1}
\end{wrapfigure}

The rapid evolution of text-to-image (T2I) generation models \citep{rombach2022high, wu2025qwen, cai2025z} has achieved unprecedented levels of visual fidelity, effectively solving the foundational challenge of raw pixel synthesis. As these models transition into ubiquitous utilities, the community's focus has naturally expanded toward accommodating an increasingly complex and diverse array of user requests \citep{feng2026gen, ye2026genclaw, he2026gems}. Modern image generation tasks are consistently pushing the boundaries of what static inference can achieve: while some prompts remain straightforward aesthetic descriptions, a rapidly growing number of contemporary requests are highly intricate, demanding external world knowledge, precise text rendering, or multi-step spatial and logical reasoning.

To tackle this escalating complexity, early efforts relied on simple test-time scaling strategies, such as prompt pre-generation refinement \citep{wang2024divide, gani2024llm} or post-generation rewriting \citep{yang2024idea2img, zhan2024prompt}. However, these homogeneous approaches remain fragile when faced with multi-constraint generation tasks. Consequently, recent research \citep{he2026gems, ren2026scope, he2026mind, chen2026genevolve} has increasingly embraced \textbf{agentic systems for image generation}, integrating advanced capabilities like knowledge retrieval, explicit reasoning, skill invocation, or iterative verification. Despite their remarkable efficacy on challenging requests, the current trajectory of agentic image generation, as shown in Figure~\ref{fig:figure1} (\textit{Left}), is hindered by two critical bottlenecks: (\textbf{\textit{i}})~\textbf{fragmentation}: existing frameworks develop specialized pipelines in silos (\textit{e.g.}, Mind-Brush \citep{he2026mind} focuses predominantly on external knowledge search), making it inherently difficult to integrate disparate capabilities or adapt their fixed topologies to diverse tasks; and (\textit{\textbf{ii}})~\textbf{compute-mismatch}: operating under a ``one-size-fits-all" paradigm, these systems impose heavily engineered workflows on every request. Consequently, even the simplest prompts are forced through costly reasoning or iterative refinement steps, unnecessarily squandering computational resources and introducing prohibitive latency. 
This leads to our pivotal research question:

\vspace{-0.4em}
\obsbox{
\textit{How can we effectively route heterogeneous prompts to their optimal workflows to balance the visual performance and computational cost?}}
\vspace{-0.1em}

To address this challenge, we first formulate \ourspace, the first unified workflow space that standardizes the execution paradigm of agentic image generation. Rather than devising yet another isolated pipeline, \ourspace systematically deconstructs the generative process into a universally applicable set of foundational primitives (\textit{e.g.}, \textit{search}, \textit{reason}, \textit{verify}, and \textit{sketch}). By strategically composing these primitives, we establish a library of standard workflow templates, ranging from a lightweight {\small``$\mathtt{DirectGen}$"} to a comprehensive {\small``$\mathtt{HybridGen}$"}. Crucially, \ourspace is not merely an exhaustive collection of existing pipelines; instead, it serves as a systematized abstraction that distills, encapsulates, and extends the core conceptual paradigms of recent fixed agentic image generation frameworks \citep{he2026gems, ren2026scope, he2026mind, chen2026genevolve, ye2026genclaw}. Ultimately, \ourspace serves as a modular and extensible foundation for the community, inherently yielding a structured space that facilitates both dynamic routing and diverse applications.

Building upon this standardized space, we further introduce \ourmethod, an experience-guided workflow router dedicated to agentic image generation. \ourmethod dynamically pairs a selected workflow template with an optimal backend image generator to mitigate the compute-mismatch bottleneck. Specifically, the routing mechanism is driven by three core components:
\ding{168}~\textbf{profiling}, extracting a lightweight task signature to quantitatively capture the intrinsic demands of the prompt;
\ding{169}~\textbf{matching}, leveraging historical execution traces and distilled experience cards to predict the overall utility of candidate plans; and \ding{171}~\textbf{filtering}, applying cost-aware Pareto filtering to dynamically prune computationally inefficient configurations.
Through this continuous cycle of routing, execution, and distillation, \ourmethod operates as a \textit{self-evolving} system, adaptively deploying a Pareto-optimal execution plan for every unique request while growing smarter over time.

To summarize, this work contributes threefold:
\vspace{-0.6em}
\begin{itemize}[leftmargin=1.6em]
    \item[\ding{182}] \textbf{Standardized Workflow Space.} We introduce \ourspace, a unified space that deconstructs agentic image generation into universal primitives and scalable templates. It establishes a modular and highly extensible infrastructure that standardizes image agentic workflows, providing a robust foundation to facilitate diverse downstream applications.
    \item[\ding{183}] \textbf{Self-Evolving Router.} We propose \ourmethod, a dynamic, experience-guided workflow router for agentic image generation. Incorporating demand profiling, utility matching, and Pareto filtering, it seamlessly pairs prompts with optimal plans and autonomously refines its precision over time via a continuous execution-distillation loop.
    \item[\ding{184}] \textbf{Empirical Validation.} Extensive experiments across diverse benchmarks demonstrate that our framework effectively mitigates both fragmentation and compute-mismatch bottlenecks. By dynamically routing prompts over \ourspace, \ourmethod achieves superior visual alignment while reducing execution costs by over $95\%$ and latency by $65\%$ compared to heavyweight static pipelines (\textit{e.g.}, GEMS). Additionally, it continuously self-evolves via accumulated experience, enabling robust generalization that boosts performance and halves computational overhead.
\end{itemize}

\section{Related Work}

\paragraph{Agentic Image Generation.} As modern image generators increasingly translate structurally detailed prompts into superior visual fidelity \citep{zhang2023text, chen2026show, wu2025qwen, cai2025z, wu2025lightgen}, agentic image generation has rapidly emerged. Early methods employed simple \textbf{test-time scaling}, coupling generators with language models for naive prompt expansion \citep{wang2024divide, gani2024llm}, chain-of-thought reasoning \citep{chen2025t2i, xiang2025promptsculptor}, or heuristic rewriting \citep{yang2024idea2img, zhan2024prompt}. While effective for standard queries, these superficial integrations struggle with intricate, multi-constraint tasks. Consequently, recent \textbf{agentic image generation} \citep{jiang2026genagent, wan2025maestro, chen2026genevolve, xu2026agentic, zhao2026toolartist, zhang2026qwen} incorporate explicit tool-use, persistent memory, and orchestration capabilities. For instance, Gen-Searcher \citep{feng2026gen} and Mind-Brush \citep{he2026mind} retrieve external multimodal evidence to resolve knowledge gaps; GenClaw \citep{ye2026genclaw} dictates spatial layouts via executable visual code; and systems like GEMS \citep{he2026gems} and SCOPE \citep{ren2026scope} orchestrate the generative lifecycle through structured skill commitments and iterative repair. However, these advanced workflows operate as isolated silos, forcing every prompt through fixed pipelines. To bridge this gap, our work abstracts these disparate capabilities into a unified foundation, termed \ourspace, and introduces \ourmethod to adaptively route each prompt to its optimal agentic configuration.

\vspace{-1.4em}
\paragraph{Agentic System Routing.} Model routing dynamically allocates queries to computational backends to balance performance and cost \citep{srivatsa2024harnessing, raschka2018model}. Existing frameworks generally follow two paradigms: (\textit{\textbf{i}})~\textbf{learning-based routing} \citep{jitkrittum2025universal, feng2025graphrouter, yue2025masrouter, zhang2025router}, which trains predictive offline policies but often struggles with shifting task distributions; and (\textit{\textbf{ii}})~\textbf{experience-based routing} \citep{zhang2026evoroute, wang2026learning, wu2025port}, which leverages real-time execution feedback for continuous refinement. While \ourmethod aligns with the experience-based paradigm, prior works focus exclusively on \textit{model-level} selection, predominantly for language models. Although recent RouteT2I \citep{xin2025adaptive} extends routing to the visual domain by dispatching requests between edge and cloud generators, it merely scales raw model capabilities within a static pipeline. Transcending simple backend allocation, we introduce the first \textit{workflow routing} framework for agentic image generation, adaptively navigating a joint space of workflow templates and multimodal generators to satisfy heterogeneous requests.

\section{Preliminary}

\paragraph{Formulation of Agentic Generation.} Let $\mathcal{X}$ and $\mathcal{Y}$ denote the spaces of textual prompts and generated images. Standard T2I generation computes $y = g(x)$, where $x \in \mathcal{X}$, $y \in \mathcal{Y}$, and $g \in \mathcal{G}$ represents a frozen generative backend (\textit{e.g.}, Qwen-Image \citep{wu2025qwen}). To accommodate intricate user requests, agentic image generation expands this static inference by introducing a foundational primitive space $\Pi = \{\pi_\mathsf{search}, \pi_\mathsf{reason}, \pi_\mathsf{sketch}, \pi_\mathsf{verify}, \dots\}$. Consequently, an agentic execution plan is mathematically formalized as a tuple $p = (w, g)$. Here, $w \in \mathcal{W}$ represents a directed topological workflow composed over $\Pi$, and $g \in \mathcal{G}$ is the paired generator. The final image is thus generated by executing the structured plan: $y = p(x)$.

\vspace{-1.4em}
\paragraph{Objective of Workflow Routing.} We define $\mathcal{P} = \mathcal{W} \times \mathcal{G}$ as the available combinatorial plan repository. Given a user prompt $x$, the objective of workflow routing is to dynamically allocate the optimal plan $p^*$ that maximizes the overall utility $U(p\mid x)$:
\setlength\abovedisplayskip{3pt}
\setlength\belowdisplayskip{3pt}
\begin{equation}
p^* = \arg\max_{p \in \mathcal{P}} U(p\mid x) = \arg\max_{p \in \mathcal{P}} \Big( S(p\mid x) - \lambda_c C(p\mid x) - \lambda_l L(p\mid x) \Big),
\label{eq:utility}
\end{equation}
where $S, C, L$ denote visual quality, execution cost, and latency, respectively, regulated by trade-off coefficients $\lambda_c$ and $\lambda_l$. 

Under this formulation, existing agentic frameworks \citep{he2026gems, ye2026genclaw} can be mathematically viewed as static instantiations where the decision space is artificially constrained to $|\mathcal{P}| = 1$ (\textit{e.g.}, consistently enforcing a fixed search \citep{he2026mind} or reasoning trajectory for every prompt \citep{ren2026scope}). Operating under such ``one-size-fits-all'' constraints inevitably leads to severe compute-mismatch. In contrast, our \ourspace establishes a structured foundation over the entire space $\mathcal{P}$, allowing \ourmethod to solve the routing objective on-the-fly and adaptively invoke compute-intensive primitives strictly when the intrinsic cognitive complexity of $x$ demands them.

\section{\ourspace: A Unified Agentic Generation Workflow Space}

\subsection{Foundational Primitives}
\begin{figure*}[!t]
\centering
\vspace{-0.4em}
\includegraphics[width=\linewidth]{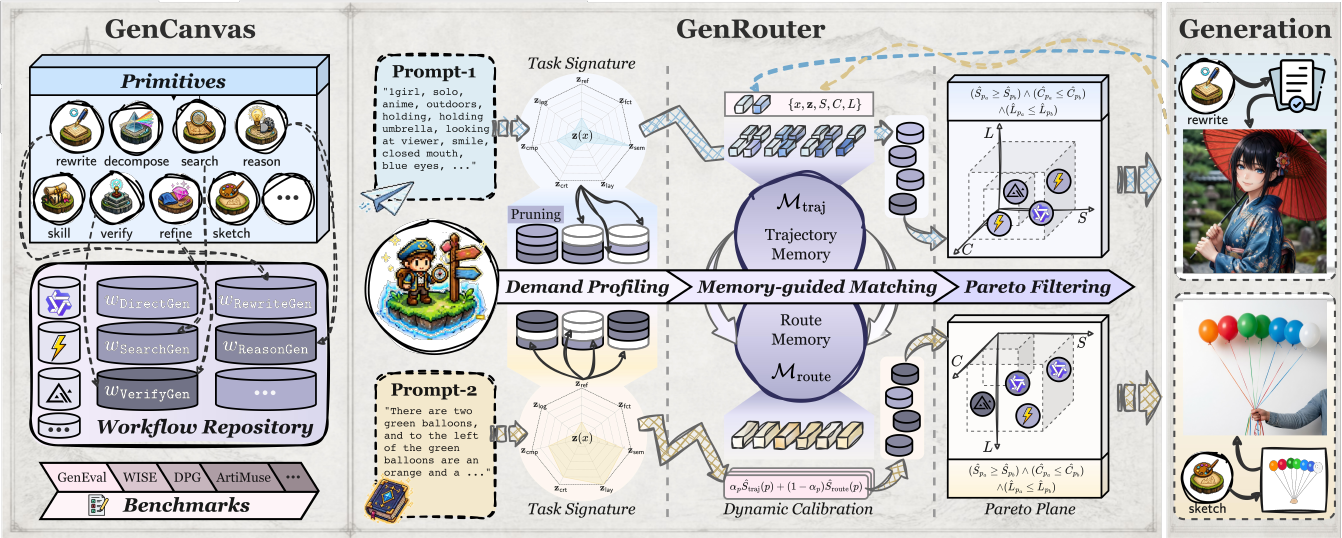}
\vspace{-1.8em}
\caption{Overview of our proposed (\textbf{\textit{Left}}) \ourspace and (\textbf{\textit{Right}}) \ourmethod.}
\label{fig:figure2}
\vspace{-0.6em}
\end{figure*}

Existing agentic image generation frameworks typically operate in isolated silos, engineering custom pipelines for specific generative sub-tasks. To resolve this fragmentation and establish an integrated foundation, we introduce \ourspace, as shown in Figure~\ref{fig:figure2} (\textit{Left}). At its core, \ourspace systematically deconstructs the complex generative process into a standardized library $\Pi$ of atomic cognitive and operational primitives. Rather than an arbitrary taxonomy, $\Pi$ is systematically distilled and extended from existing agentic paradigms (\textit{e.g.}, GEMS \citep{he2026gems}, GenClaw \citep{ye2026genclaw}) to ensure comprehensive coverage. Formally, we define this foundational space with eight primitives:
\begin{equation}
    \Pi = \{\pi_\mathsf{rewrite}, \pi_\mathsf{decompose}, \pi_\mathsf{search}, \pi_\mathsf{reason}, \pi_\mathsf{skill}, \pi_\mathsf{verify}, \pi_\mathsf{refine}, \pi_\mathsf{sketch}\}.
\end{equation}
Functioning as cohesive building blocks that abstract away underlying model mechanics, these primitives encapsulate distinct capabilities, ranging from external knowledge retrieval ($\pi_\mathsf{search}$) and logical inference ($\pi_\mathsf{reason}$) to spatial layout compilation ($\pi_\mathsf{sketch}$), providing the essential modular infrastructure to construct diverse agentic workflows.

\subsection{Workflow Design and Taxonomy}

While the primitive library $\Pi$ theoretically enables arbitrary combinations, unconstrained tool invocation often causes execution instability and redundant overhead. To standardize generation and bound the execution space ($\mathcal{P} = \mathcal{W} \times \mathcal{G}$), \ourspace decouples topological execution logic from the terminal generator $g \in \mathcal{G}$ by predefining a set of reliable workflow templates $w \in \mathcal{W}$. Inspired by diverse cognitive demands, these templates are categorized into four hierarchical levels:

\vspace{-0.6em}
\begin{itemize}[leftmargin=1.6em]
    \item[\ding{111}] \textbf{Semantic Alignment (}$w_\mathtt{DirectGen}, w_\mathtt{RewriteGen}$\textbf{).} For aesthetic or straightforward queries, these templates bypass heavy external tools, either directly invoking $g$ or employing $\pi_\mathsf{rewrite}$ to structurally enrich underspecified inputs before generation.
    \item[\ding{111}] \textbf{External Grounding (}$w_\mathtt{SearchGen}, w_\mathtt{RefGen}$\textbf{).} To mitigate factual or structural hallucinations regarding real-world and long-tail entities, these workflows explicitly invoke $\pi_\mathsf{search}$ to retrieve external textual knowledge or visual references prior to execution.
    \item[\ding{111}] \textbf{Structural Reasoning (}$w_\mathtt{ReasonGen}, w_\mathtt{SkillGen}, w_\mathtt{SketchGen}$\textbf{).}
    For demanding spatial layouts or precise constraints, these templates enforce rigorous pre-generation planning, leveraging $\pi_\mathsf{reason}$ for logical deduction, $\pi_\mathsf{skill}$ for specialized formatting, or $\pi_\mathsf{sketch}$ to compile structural intents into visual layout codes.
    \item[\ding{111}] \textbf{Iterative Refinement (}$w_\mathtt{VerifyGen}, w_\mathtt{HybridGen}$\textbf{).}
    To handle multi-constraint prompts prone to one-shot failures, these workflows construct closed-loop structures via $\pi_\mathsf{decompose}$ and $\pi_\mathsf{verify}$. A localized memory traces multimodal feedback, systematically guiding $\pi_\mathsf{refine}$ to correct misalignments until convergence.
\end{itemize}
\vspace{-0.6em}
By formalizing these templates, \ourspace establishes a robust and standardized infrastructure for deploying diverse generative strategies. Detailed configurations are provided in Appendix~\S\ref{app:gencanvas_implementation}.

\subsection{\ourspace Codebase}

Beyond a conceptual taxonomy, \ourspace serves as a highly extensible, open-source codebase dedicated to the construction and evaluation of agentic image generation workflows.

\vspace{-1.4em}
\paragraph{Implementation.} Engineered with strict modularity, the framework cleanly decouples topological configurations from model backends. This plug-and-play architecture enables the flexible substitution of base models for individual primitives and terminal generators $g \in \mathcal{G}$ (\textit{e.g.}, Qwen-Image \citep{wu2025qwen} or Z-Image \citep{cai2025z}), while seamlessly supporting the integration of new primitives and customized workflows. Furthermore, \ourspace logically unifies recent paradigms, mapping their execution paths to specific instantiations within our taxonomy, as summarized in Table~\ref{tab:workflow_mapping}.

\vspace{-1.4em}
\paragraph{Evaluation.} To facilitate comprehensive assessments, \ourspace provides out-of-the-box support for mainstream evaluation suites (\textit{e.g.}, GenEval \citep{ghosh2023geneval}, DPG-Bench \citep{hu2024ella}, WISE \citep{niu2025wise}) and downstream benchmarks (\textit{e.g.}, LongText-Bench \citep{geng2025x}, SpatialGenEval \citep{wang2026everything}, ArtiMuse \citep{cao2026artimuse}). By standardizing input-output interfaces across all templates, the engine natively tracks execution latency and token consumption, offering a rigorous testbed for evaluating complex generation pipelines.

\begingroup
\setlength{\tabcolsep}{6pt} 
\begin{table*}[!t]
\renewcommand{\arraystretch}{1.25} 
  \centering
  \caption{Core workflow templates and topological paths in \ourspace. We abstract diverse generative demands into $9$ structured templates over $\Pi$. Representative methods are mapped solely for conceptual alignment, not strict implementation equivalence. Icons classify these prior works into test-time scaling (\raisebox{-0.15em}{\includegraphics[height=0.7em]{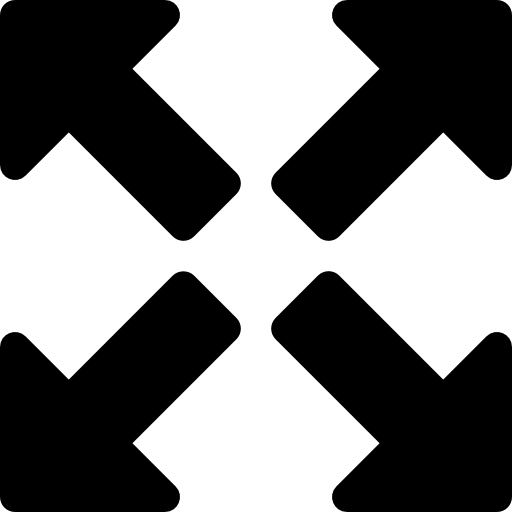}}) or agentic workflow (\raisebox{-0.15em}{\includegraphics[height=0.8em]{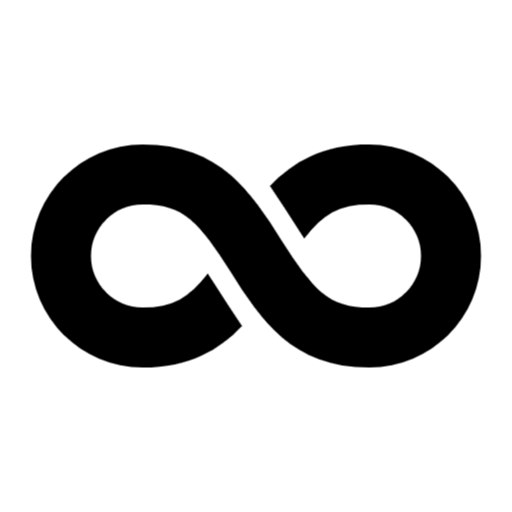}}) for image generation.}
  \vspace{-0.9em}
  \resizebox{\linewidth}{!}{
   \begin{tabular}{llcl}
    \hlineB{2.5}
    \rowcolor{CadetBlue!20} 
    \textbf{Cognitive Level} & \textbf{Workflow ($w \in \mathcal{W}$)} & \textbf{Topological Template} & \textbf{Representative Method} \\
    \hlineB{1.5}
    
    \textbf{I: Semantic Alignment} 
    & $\mathtt{DirectGen}$ & $g$ & \\
    & $\mathtt{RewriteGen}$ & $\pi_\mathsf{rewrite} \rightarrow g$ & \raisebox{-0.15em}{\includegraphics[height=0.8em]{figures/expand2.png}}~BeautifulPrompt \citep{cao2023beautifulprompt} \\
    
    \rowcolor{gray!10}
    \textbf{II: External Grounding} 
    & $\mathtt{SearchGen}$ & $\pi_\mathsf{search}\text{ (text)} \rightarrow \pi_\mathsf{rewrite} \rightarrow g$ & \raisebox{-0.15em}{\includegraphics[height=0.9em]{figures/loop2.png}}~Gen-Searcher \citep{feng2026gen} \\
    \rowcolor{gray!10}
    & $\mathtt{RefGen}$ & $\pi_\mathsf{search}\text{ (visual)} \rightarrow \pi_\mathsf{rewrite} \rightarrow g$ & \raisebox{-0.15em}{\includegraphics[height=0.9em]{figures/loop2.png}}~Mind-Brush \citep{he2026mind} \\
    
    \textbf{III: Structural Reasoning} 
    & $\mathtt{ReasonGen}$ & $\pi_\mathsf{reason} \rightarrow \pi_\mathsf{rewrite} \rightarrow g$ & \raisebox{-0.15em}{\includegraphics[height=0.8em]{figures/expand2.png}}~Self-CoT \citep{deng2025emerging} \\
    & $\mathtt{SkillGen}$ & $\pi_\mathsf{skill} \rightarrow \pi_\mathsf{rewrite} \rightarrow g$ & \raisebox{-0.15em}{\includegraphics[height=0.9em]{figures/loop2.png}}~GEMS \citep{he2026gems} \\
    & $\mathtt{SketchGen}$ & $\pi_\mathsf{sketch} \rightarrow g$ & \raisebox{-0.15em}{\includegraphics[height=0.9em]{figures/loop2.png}}~GenClaw \citep{ye2026genclaw} \\
    
    \rowcolor{gray!10}
    \textbf{IV: Iterative Refinement} 
    & $\mathtt{VerifyGen}$ & $\pi_\mathsf{decompose} \rightarrow \mathbf{Loop}(g \rightarrow \pi_\mathsf{verify} \rightarrow \pi_\mathsf{refine})$ & \raisebox{-0.15em}{\includegraphics[height=0.9em]{figures/loop2.png}}~GEMS \citep{he2026gems} \\
    \rowcolor{gray!10}
    & $\mathtt{HybridGen}$ & $\{\pi_\mathsf{search}, \pi_\mathsf{sketch}\} \rightarrow \mathbf{Loop}(g \rightarrow \pi_\mathsf{verify} \rightarrow \pi_\mathsf{refine})$ & \raisebox{-0.15em}{\includegraphics[height=0.9em]{figures/loop2.png}}~SCOPE \citep{ren2026scope} \\
    
    \hlineB{2.5}
   \end{tabular}
  }
  \label{tab:workflow_mapping}
  \vspace{-0.6em}
\end{table*} 
\endgroup

\section{\ourmethod: A Self-Evolving Agentic Image Workflow Router}

While \ourspace provides a unified abstraction for agentic templates, practical deployment requires dynamically matching incoming prompts with appropriate workflows. Fixed pipelines inevitably cause compute-mismatch, and manual selection is unscalable. To bridge this gap, we propose \ourmethod, as shown in Figure~\ref{fig:figure2} (\textit{Right}), an adaptive router that selects an optimal workflow-generator plan $p=(w,g)$ to balance visual quality and computational efficiency.

Given a prompt $x$, \ourmethod estimates the expected utility $\hat{U}(p\mid x)$, as defined in Eq.~(\ref{eq:utility}), for each candidate plan. To ensure the search space remains empirically comparable, auxiliary primitive backends (\textit{e.g.}, search engines and verifiers) remain modular but fixed under a given routing configuration. The final routing objective is formalized as:
\begin{equation}
\hat{p} = \arg\max_{p\in\mathcal{P}_\text{pareto}(x)} \hat{U}(p\mid x),
\end{equation}
where $\mathcal{P}_\text{pareto}(x)$ represents the subset of plans that satisfy capability and task-compatibility constraints, naturally allocating advanced workflows only to demanding requests.

\subsection{Demand Profiling}
\label{sec:5.1}

Conventional LLM (agent) routing often relies on semantic similarity. However, T2I generation is highly sensitive to subtle intent shifts; minor modifications (\textit{e.g.}, adding \textit{``with the exact text `ICLR 2027'''}) can drastically alter generative demands despite minimal embedding distance. Furthermore, direct LLM-as-router also suffers from severe calibration issues and hallucination. Instead, \ourmethod utilizes a lightweight LLM (\textit{e.g.}, Qwen3.5-4B \citep{yang2025qwen3}) strictly as a \textit{profiler} to extract intrinsic cognitive demands of the prompt, guiding subsequent candidate construction.

\vspace{-1.4em}
\paragraph{Task Signature Extraction.} The profiler first maps $x$ into a seven-dimensional intent signature:
\begin{equation}
\mathbf{z}(x) = (\mathbf{z}_\mathsf{sem}, \mathbf{z}_\mathsf{fct}, \mathbf{z}_\mathsf{ref}, \mathbf{z}_\mathsf{log}, \mathbf{z}_\mathsf{cmp}, \mathbf{z}_\mathsf{crt}, \mathbf{z}_\mathsf{lay}) \in \{0, 1, 2, 3, 4, 5\}^7.
\end{equation}
These axes represent semantic articulation, factual grounding, visual referencing, logical deduction, compositional heuristics, evaluative critique, and spatial layout, respectively. Scores $\ge 3$ indicate high-need thresholds. This decoupling offers three benefits:
\ding{192} \textit{interpretability} through tracing decisions to specific needs; \ding{193} \textit{calibration} by treating the signature as a prior rather than a rigid path; and \ding{194} \textit{evolvability} via real-world execution feedback (detailed in Section~\S\ref{sec:memory_matching}).

\vspace{-1.4em}
\paragraph{Candidate Pruning.}  To avoid evaluating the entire space $\mathcal{P}$, \ourmethod derives a valid candidate set $\mathcal{P}_\text{valid}(x)$ via two pruning mechanisms:
\ding{182} \textit{capability compatibility}: a plan $(w, g)$ is invalid if $w$ requires capabilities $g$ lacks (\textit{e.g.}, $w_\mathtt{RefGen}$ requires visual conditioning); and
\ding{183} \textit{signature gating}: to prevent compute-mismatch, heavyweight templates are explicitly gated by $\mathbf{z}(x)$. For instance, $w_\mathtt{HybridGen}$ is activated only for complex prompts:
\begin{equation}
    \text{active}(w_\mathtt{HybridGen}) = \mathbb{I} \left[ \sum_{k \in \mathcal{H}} \mathbb{I}(\mathbf{z}_k \ge 3) \ge 2 \right], \quad \mathcal{H} = \{\mathbf{z}_\mathsf{fct}, \mathbf{z}_\mathsf{ref}, \mathbf{z}_\mathsf{log}, \mathbf{z}_\mathsf{cmp}, \mathbf{z}_\mathsf{crt}, \mathbf{z}_\mathsf{lay}\}.
\end{equation}
Cascading these constraints aggressively distills the combinatorial space, setting the stage for memory-guided utility estimation.

\subsection{Memory-Guided Matching}
\label{sec:memory_matching}

While the task signature $\mathbf{z}(x)$ identifies fundamental generative needs, it acts as a static prior oblivious to environment-specific performance variability. To adapt dynamically, \ourmethod maintains a dual-memory system: \textit{trajectory memory} and \textit{route memory}.

\vspace{-1.4em}
\paragraph{Dual-Memory Experience.} Trajectory memory $\mathcal{M}_\text{traj}$ logs instance-level execution records. For a candidate plan $p$, it retrieves the top-$k$ most similar historical prompts and estimates the expected utility $\hat{U}_\text{traj}(p)$ by aggregating outcomes exclusively from past instances that utilized the exact same plan $p$. To mitigate data sparsity when exact matches are unavailable, route memory $\mathcal{M}_\text{route}$ periodically distills these trajectory records into a bucket-level statistical representation. By grouping historical executions into coarse task categories based on their signatures, $\mathcal{M}_\text{route}$ provides robust, aggregated priors (\textit{e.g.}, mean quality, cost, and latency) for each plan.

\vspace{-1.4em}
\paragraph{Dynamic Calibration.}
To determine the final plan, \ourmethod fuses the estimates from both memories. The expected quality (and analogously, cost and latency) is computed as a confidence-weighted sum:
\begin{equation}
\hat{S}_p = \alpha_p \hat{S}_\text{traj}(p) + (1-\alpha_p) \hat{S}_\text{route}(p),
\end{equation}
where the confidence parameter $\alpha_p$ increases proportionally with the number of retrieved trajectory records containing $p$. If empirical memory is entirely absent (\textit{e.g.}, during cold-start), the system falls back to a deterministic threshold prior derived directly from $\mathbf{z}(x)$. This architecture ensures that \ourmethod seamlessly transitions from signature-based priors to robust empirical routing, autonomously self-correcting based on realized compute costs and visual outcomes.

\subsection{Pareto Filtering}

After deriving the metric estimates $\{\hat{S}_p,\hat{C}_p,\hat{L}_p\}$ for each $p\in\mathcal{P}_\text{valid}(x)$, relying solely on the scalarized utility $\hat{U}_p$ can be vulnerable to anomalous selections, where a strictly inferior plan might be chosen due to linear weighting artifacts. To achieve more robust routing, we introduce a Pareto filtering step prior to final selection. A plan $p_a$ dominates $p_b$ if it is no worse across all dimensions and strictly better in at least one:
\begin{equation}
(\hat{S}_{p_a} \ge \hat{S}_{p_b}) \land (\hat{C}_{p_a} \le \hat{C}_{p_b}) \land (\hat{L}_{p_a} \le \hat{L}_{p_b}), \quad p_a \neq p_b.
\end{equation}
Defining the non-dominated Pareto set as $\mathcal{P}_\text{pareto}$, we deterministically prune suboptimal configurations that offer no distinct trade-off advantages. The final plan is then selected as $\hat{p}=\arg\max_{p\in\mathcal{P}_\text{pareto}}\hat{U}_p$.

Through this multi-stage refinement, \textit{i.e.}, \ding{168} \textbf{demand profiling}, \ding{169} \textbf{memory-guided matching}, and \ding{171} \textbf{Pareto filtering}, \ourmethod realizes a self-evolving loop: the system continuously accumulates empirical records to refine its utility landscape. This enables \ourmethod to adaptively bridge the gap between initial signature-based priors and the nuanced performance characteristics of real-world deployment, ensuring long-term optimization without manual retraining.

\section{Experiments}

In this section, we conduct extensive experiments to answer the following key research questions:
(\textbf{RQ1})~Can \ourmethod effectively navigate \ourspace to achieve a superior trade-off between quality, cost, and latency compared to static agentic pipelines?
(\textbf{RQ2})~Are diverse workflow templates of \ourspace necessary, and does \ourmethod make interpretable routing decisions?
(\textbf{RQ3})~Can \ourmethod continuously self-evolve via experience accumulation and generalize effectively?

\subsection{Experimental Settings}

\paragraph{Baselines.} We benchmark against direct generators and three open-sourced agentic image generation workflows: Mind-Brush \citep{he2026mind} (with search), SCOPE \citep{ren2026scope} (with hybrid skills), and GEMS \citep{he2026gems} (with verification). For fair comparison, we unify all underlying LLM/MLLM engines while preserving their original workflow topologies.

\vspace{-1.4em}
\paragraph{Benchmarks and Metrics.} We evaluate \ourmethod across two categories: \textbf{(I)~mainstream} benchmarks featuring broad prompt distributions, \textit{i.e.}, WISE \citep{niu2025wise}, DPG-Bench \citep{hu2024ella}, OneIG-Bench \citep{changoneig}, GenEval2 \citep{kamath2025geneval}; and \textbf{(II)~downstream} benchmarks dedicated to targeted capability analysis, \textit{i.e.}, LongText \citep{geng2025x}, SpatialGenEval \citep{wang2026everything}, ArtiMuse \citep{cao2026artimuse}.
For metrics, we evaluate \ourmethod across three core dimensions: \ding{182}~\textit{performance}: official benchmark scores; 
\ding{183}~\textit{cost}: exact token and API consumption; and \ding{184}~\textit{latency}: end-to-end workflow execution time. To ensure fair cross-workflow comparisons, generator-specific inference latency is explicitly isolated.

\vspace{-1.4em}
\paragraph{Generative Backends.} Our experiments primarily utilize two advanced image generators to instantiate the terminal action space ($\mathcal{G}$): \raisebox{-0.15em}{\includegraphics[height=1em]{figures/z-icon4.png}}~Z-Image-Turbo \citep{cai2025z} and \raisebox{-0.15em}{\includegraphics[height=1em]{figures/qwen_logo.png}}~Qwen-Image-2512 \citep{wu2025qwen}. To support reference-dependent workflows (\textit{e.g.}, $w_{\mathtt{RefGen}}$ and $w_{\mathtt{SketchGen}}$), we integrate Qwen-Image-Edit-2511 as a visual-conditioning mode for the Qwen-Image family.

\begin{table*}[t]
\centering
\caption{Performance comparison on WISE, DPG-Bench and GenEval2 benchmarks. The \textbf{best} and \underline{second best} results are highlighted.}
\vspace{-0.8em}
\renewcommand{\arraystretch}{1.3}
\resizebox{\textwidth}{!}{%
\begin{tabular}{cc|ccccccccc}
\hlineB{2.5}
\multirow{2}{*}{\textbf{Generator}} & \multirow{2}{*}{\textbf{Method}} & \multicolumn{3}{c}{\textbf{WISE}} & \multicolumn{3}{c}{\textbf{DPG-Bench}} & \multicolumn{3}{c}{\textbf{GenEval2}} \\
\cmidrule(lr){3-5} \cmidrule(lr){6-8} \cmidrule(lr){9-11} 
& & \textbf{Perf. (\%)} & \textbf{Cost (\$)} & \textbf{Latency (h)} & \textbf{Perf. (\%)} & \textbf{Cost (\$)} & \textbf{Latency (h)} & \textbf{Perf. (\%)} & \textbf{Cost (\$)} & \textbf{Latency (h)} \\
\hlineB{2}
\multirow{5}{*}{\rotatebox{90}{\qwenlogo~Qwen-Image~}} 
& Original \citep{wu2025qwen} & 0.53 & 0.00 & 0.00 & 87.21 & 0.00 & 0.00 & 29.0 & 0.00 & 0.00 \\
& Mind-Brush \citep{he2026mind} & 0.68 & 4.40 & \textbf{3.65} & 83.74 & \underline{2.42} & \underline{2.71} & 32.1 & \textbf{1.80} & \textbf{2.38} \\
& SCOPE \citep{ren2026scope} & \textbf{0.88} & \underline{1.97} & 4.34 & \underline{86.94} & 4.88 & 9.42 & 46.3 & \underline{1.81} & 3.89 \\
& GEMS \citep{he2026gems} & \underline{0.80} & 38.17 & 11.94 & 85.59 & 60.31 & 15.23 & \underline{70.4} & 40.43 & 8.43 \\
& \cellcolor{blue!10} \ourspace\textbf{+} \ourmethod & \cellcolor{blue!10}\textbf{0.88} & \cellcolor{blue!10}\textbf{1.57} & \cellcolor{blue!10}\underline{4.16} & \cellcolor{blue!10}\textbf{87.39} & \cellcolor{blue!10}\textbf{1.41} & \cellcolor{blue!10}\textbf{2.52} & \cellcolor{blue!10}\textbf{71.6} & \cellcolor{blue!10}2.03 & \cellcolor{blue!10}\underline{3.59} \\
\hline
\multirow{5}{*}{\rotatebox{90}{\zlogo~Z-Image~}} 
& Original \citep{cai2025z} & 0.57 & 0.00 & 0.00 & 85.01 & 0.00 & 0.00 & 31.0 & 0.00 & 0.00 \\
& Mind-Brush \citep{he2026mind} & 0.58 & 4.20 & \underline{3.44} & 85.16 & \underline{2.44} & \underline{1.28} & 30.1 & \underline{1.80} & \textbf{2.55} \\
& SCOPE \citep{ren2026scope} & \underline{0.85} & \underline{1.93} & 4.26 & 85.03 & 4.86 & 9.65 & 40.0 & 1.82 & 4.33 \\
& GEMS \citep{he2026gems} & 0.81 & 24.63 & 11.83 & \textbf{86.01} & 46.33 & 18.69 & \textbf{63.5} & 19.63 & 8.32 \\
& \cellcolor{Goldenrod!30} \ourspace\textbf{+} \ourmethod & \cellcolor{Goldenrod!30}\textbf{0.86} & \cellcolor{Goldenrod!30}\textbf{1.33} & \cellcolor{Goldenrod!30}\textbf{2.00} & \cellcolor{Goldenrod!30}\underline{85.49} & \cellcolor{Goldenrod!30}\textbf{1.00} & \cellcolor{Goldenrod!30}\textbf{1.13} & \cellcolor{Goldenrod!30}\underline{57.6} & \cellcolor{Goldenrod!30}\textbf{1.76} & \cellcolor{Goldenrod!30}\underline{3.85} \\
\hline
\qwenlogo~\textbf{+}~\zlogo & \cellcolor{mygrey!10} \ourspace\textbf{+} \ourmethod & \cellcolor{mygrey!10}0.87 & \cellcolor{mygrey!10}1.50 & \cellcolor{mygrey!10}3.16 & \cellcolor{mygrey!10}86.44 & \cellcolor{mygrey!10}1.17 & \cellcolor{mygrey!10}1.85 & \cellcolor{mygrey!10}67.5 & \cellcolor{mygrey!10}1.87 & \cellcolor{mygrey!10}3.76 \\
\hlineB{2.5}
\end{tabular}%
}
\label{tab:main_comp_1}
\end{table*}

\begin{table*}[t]
\centering
\vspace{-0.4em}
\caption{Performance comparison on OneIG-Bench benchmarks. ``Average" includes five benchmarks' results in both Tables~\ref{tab:main_comp_1} and~\ref{tab:main_comp_2}. }
\vspace{-0.8em}
\renewcommand{\arraystretch}{1.3}
\resizebox{\textwidth}{!}{%
\begin{tabular}{cc|cccccc|ccc}
\hlineB{2.5}
\multirow{2}{*}{\textbf{Generator}} & \multirow{2}{*}{\textbf{Method}} & \multicolumn{3}{c}{\textbf{OneIG-EN}} & \multicolumn{3}{c}{\textbf{OneIG-CN}} & \multicolumn{3}{c}{\textbf{Average}} \\
\cmidrule(lr){3-5} \cmidrule(lr){6-8} \cmidrule(lr){9-11} 
& & \textbf{Perf. (\%)} & \textbf{Cost (\$)} & \textbf{Latency (h)} & \textbf{Perf. (\%)} & \textbf{Cost (\$)} & \textbf{Latency (h)} & \textbf{Perf. (\%)} & \textbf{Cost (\$)} & \textbf{Latency (h)} \\
\hlineB{2}
\multirow{5}{*}{\rotatebox{90}{\qwenlogo~Qwen-Image~}} 
& Original \citep{wu2025qwen} & 0.487 & 0.00 & 0.00 & 0.489 & 0.00 & 0.00 & 53.4 & 0.00 & 0.00 \\
& Mind-Brush \citep{he2026mind} & 0.508 & \textbf{3.30} & \textbf{4.95} & 0.516 & \textbf{4.34} & \underline{7.75} & 57.2 & \underline{3.25} & \textbf{4.29} \\
& SCOPE \citep{ren2026scope} & 0.521 & \underline{4.30} & 8.87 & 0.504 & \underline{4.92} & 10.39 & 64.7 & 3.58 & 7.38 \\
& GEMS \citep{he2026gems} & \underline{0.542} & 73.70 & 16.21 & \underline{0.532} & 86.90 & 16.27 & \underline{68.7} & 59.70 & 13.62 \\
& \cellcolor{blue!10} \ourspace\textbf{+} \ourmethod & \cellcolor{blue!10}\textbf{0.553} & \cellcolor{blue!10}5.50 & \cellcolor{blue!10}\underline{5.92} & \cellcolor{blue!10}\textbf{0.541} & \cellcolor{blue!10}\textbf{4.34} & \cellcolor{blue!10}\textbf{7.23} & \cellcolor{blue!10}\textbf{71.3} & \cellcolor{blue!10}\textbf{2.97} & \cellcolor{blue!10}\underline{4.68} \\
\hline
\multirow{5}{*}{\rotatebox{90}{\zlogo~Z-Image~}} 
& Original \citep{cai2025z} & 0.526 & 0.00 & 0.00 & 0.501 & 0.00 & 0.00 & 55.1 & 0.00 & 0.00 \\
& Mind-Brush \citep{he2026mind} & 0.517 & \textbf{3.45} & \textbf{4.99} & 0.489 & \underline{4.63} & \underline{8.00} & 54.8 & \underline{3.30} & \textbf{4.05} \\
& SCOPE \citep{ren2026scope} & 0.523 & \underline{4.18} & 7.74 & 0.511 & 4.71 & 8.95 & 62.7 & 3.50 & 6.99 \\
& GEMS \citep{he2026gems} & \underline{0.569} & 43.13 & 14.50 & \textbf{0.552} & 53.37 & 16.72 & \textbf{68.5} & 37.42 & 14.01 \\
& \cellcolor{Goldenrod!30} \ourspace\textbf{+} \ourmethod & \cellcolor{Goldenrod!30}\textbf{0.570} & \cellcolor{Goldenrod!30}6.67 & \cellcolor{Goldenrod!30}\underline{5.95} & \cellcolor{Goldenrod!30}\underline{0.545} & \cellcolor{Goldenrod!30}\textbf{4.56} & \cellcolor{Goldenrod!30}\textbf{7.62} & \cellcolor{Goldenrod!30}\underline{68.1} & \cellcolor{Goldenrod!30}\textbf{3.06} & \cellcolor{Goldenrod!30}\underline{4.11} \\
\hline
\qwenlogo~\textbf{+}~\zlogo & \cellcolor{mygrey!10} \ourspace\textbf{+} \ourmethod & \cellcolor{mygrey!10}0.561 & \cellcolor{mygrey!10}6.12 & \cellcolor{mygrey!10}5.94 & \cellcolor{mygrey!10}0.542 & \cellcolor{mygrey!10}4.44 & \cellcolor{mygrey!10}7.41 & \cellcolor{mygrey!10}70.2 & \cellcolor{mygrey!10}3.02 & \cellcolor{mygrey!10}4.42 \\
\hlineB{2.5}
\end{tabular}%
}
\label{tab:main_comp_2}
\vspace{-0.6em}
\end{table*}

\vspace{-1.4em}
\paragraph{Method Configurations.} The demand profiler leverages a lightweight local LLM (Qwen3.5-4B \citep{yang2025qwen3}). While our primitive backends are designed to be entirely plug-and-play, we instantiate all general-purpose linguistic and visual primitives (\textit{e.g.}, $\mathsf{reason}, \mathsf{verify}$) using Kimi K2.5 \citep{team2026kimi} (following GEMS \citep{he2026gems}) to rigorously isolate the performance gains of our routing mechanism. External grounding relies on the Serper Search\footnote{https://serper.dev/}. For the routing utility, $\lambda_c$ and $\lambda_l$ are set to $5.0$ and $0.0006$, respectively. More configurations are detailed in Appendix~\S\ref{app:gencanvas_implementation}.


\subsection{Main Results}

\begin{figure*}[!t]
\centering
\includegraphics[width=\linewidth]{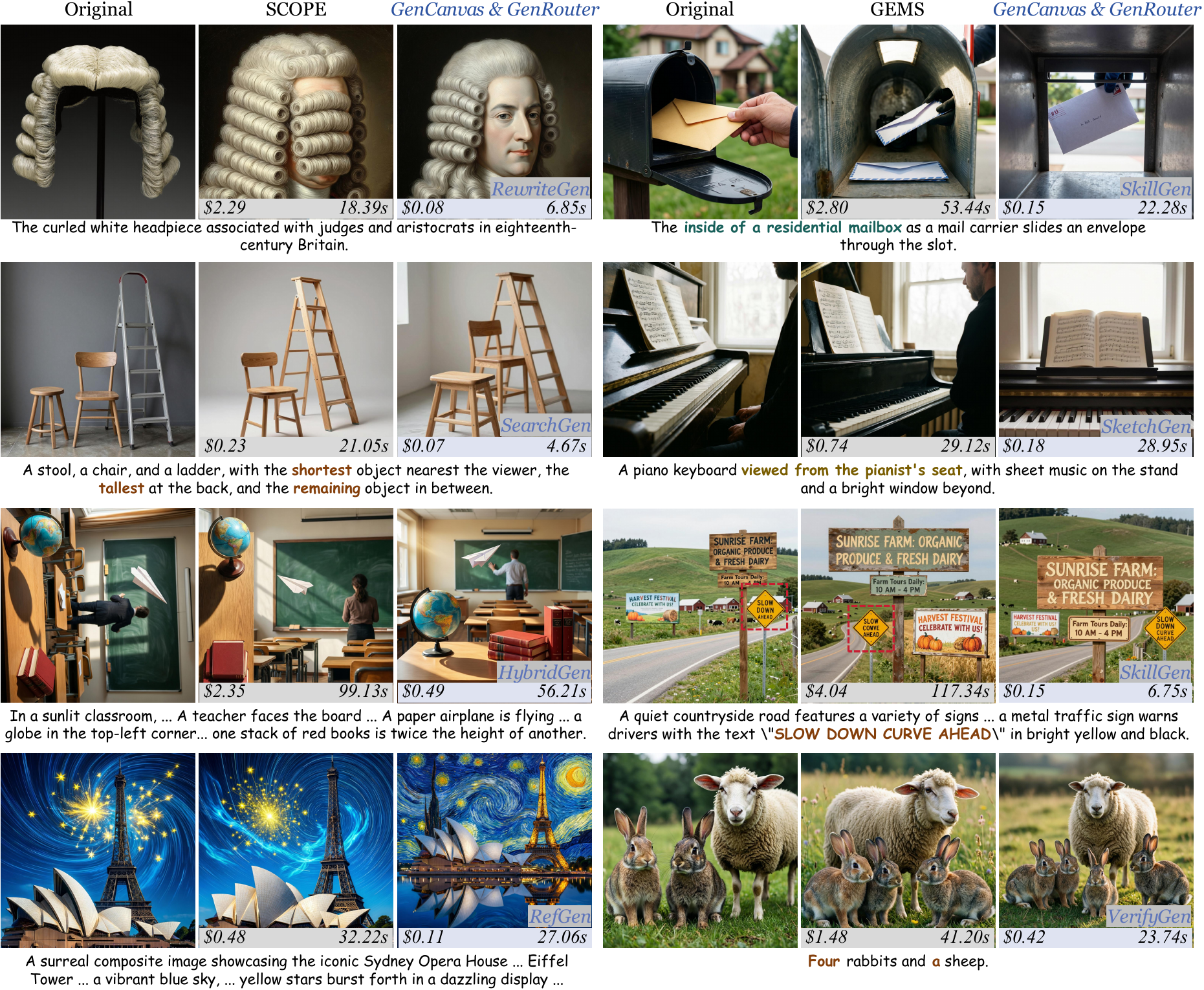}
\vspace{-1.6em}
\caption{Qualitative comparison. Left side shows cost (scaled by $0.01$); right side shows latency.}
\label{fig:qualitative}
\vspace{-0.6em}
\end{figure*}

To answer \textbf{RQ1}, we quantitatively evaluate the generation performance, API cost, and execution latency of our \ourmethod within \ourspace against static agentic baselines, summarized in Tables~\ref{tab:main_comp_1} and~\ref{tab:main_comp_2}, alongside qualitative comparisons in Figures~\ref{fig:teaser}, \ref{fig:qualitative}, \ref{fig:exhibition} and~\ref{fig:exhibition_2}. Key observations are summarized:

\textbf{Obs.\ding{182} \ourmethod significantly outperforms static pipelines while requiring a fraction of their computational overhead.}
As shown in the average results across five benchmarks in Table~\ref{tab:main_comp_2}, \ourmethod operating over \ourspace with \qwenlogo~achieves the highest overall performance ($71.3\%$) while reducing execution cost by over $95\%$ ($\$2.97$ \textit{vs.} $\$59.70$) and latency by $65\%$ ($4.68$h \textit{vs.} $13.62$h) compared to the GEMS framework. Similar efficiency gains are consistently observed with the \zlogo~backend. Even when compared to lighter orchestration pipelines like SCOPE, the synergy of \ourspace and \ourmethod consistently delivers superior visual alignment with strictly lower overhead. This confirms that dynamically routing prompts to tailored workflows mitigates the compute-mismatch bottleneck inherent in ``one-size-fits-all" systems.

\textbf{Obs.\ding{183} Dynamic routing effectively mitigates visual failure modes by allocating task-specific workflow templates.}
Qualitative comparisons in Figures~\ref{fig:qualitative},~\ref{fig:exhibition} and~\ref{fig:exhibition_2} illustrate that static pipelines often misallocate resources, either failing to resolve complex spatial constraints due to insufficient reasoning, or over-complicating simple queries. By adaptively dispatching structurally precise templates from \ourspace (\textit{e.g.}, invoking $w_{\mathtt{SketchGen}}$ for multi-object layouts or $w_{\mathtt{SkillGen}}$ for text rendering), \ourmethod successfully rectifies these errors while dramatically reducing per-prompt latency and cost (\textit{e.g.}, $\$0.18$ \textit{vs.} $\$0.74$ in the piano case). Furthermore, the diverse high-fidelity generations presented in Figure~\ref{fig:teaser} demonstrate the combined framework's robust versatility across various artistic styles, long-tail entities, and intricate semantic demands.



\subsection{Routing Interpretability Analysis}

\begin{figure*}[!t]
\centering
\includegraphics[width=\linewidth]{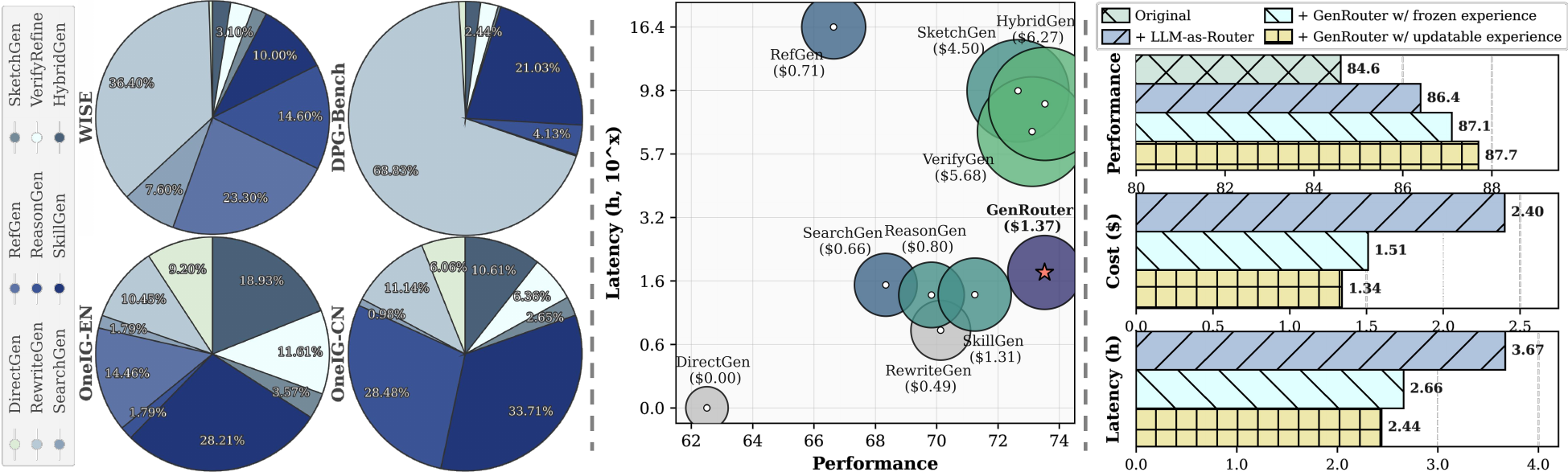}
\vspace{-1.6em} 
\caption{(\textbf{\textit{Left}}) Routing distribution of \ourmethod; (\textbf{\textit{Middle}}) Utility comparison between isolated \ourspace workflows; (\textbf{\textit{Right}}) Zero-shot generalization on DPG-Bench with WISE experience.}
\label{fig:figure3}
\vspace{-0.6em}
\end{figure*}

To answer \textbf{RQ2}, we investigate the routing behaviors of \ourmethod within \ourspace. We analyze the workflow distributions on benchmarks from Tables~\ref{tab:main_comp_1} and~\ref{tab:main_comp_2} in Figure~\ref{fig:figure3} (\textit{Left}). Furthermore, for comprehensive evaluation across diverse scenarios, we construct a \textbf{500-prompt mixed test set} sampling from nine mainstream and downstream benchmarks (details in Appendix~\S\ref{app:c.1}). We compare the performance of \ourmethod against single-workflow execution in \ourspace on this mixed set in Figure~\ref{fig:figure3} (\textit{Middle}). Our key observations are summarized below:

\textbf{Obs.\ding{184} Routing distributions strongly correlate with the intrinsic demands of heterogeneous benchmarks, proving the necessity of diverse templates.}
As shown in Figure~\ref{fig:figure3} (\textit{Left}), \ourmethod exhibits distinct routing preferences tailored to each benchmark's characteristics. For instance, the aesthetics-focused DPG-Bench is dominated by lightweight text enhancements like $w_{\mathtt{RewriteGen}}$ ($68.83\%$) and $w_{\mathtt{SkillGen}}$ ($21.03\%$), with minimal need for heavy verification. Conversely, prompts from OneIG-EN and OneIG-CN, which feature complex spatial and logical constraints, trigger a much broader utilization of structurally demanding templates, notably $w_{\mathtt{HybridGen}}$ ($18.93\%$ and $10.61\%$) and $w_{\mathtt{ReasonGen}}$ ($28.48\%$ for OneIG-CN). This dynamic adaptation confirms that a diverse workflow repository is necessary to optimally resolve heterogeneous generative intents.

\textbf{Obs.\ding{185} \ourmethod achieves Pareto-optimal efficiency by allocating compute-intensive workflows strictly to complex queries.}
Figure~\ref{fig:figure3} (\textit{Middle}) plots the utility landscape of all isolated workflow templates in \ourspace on the mixed test set. While comprehensive templates like $w_{\mathtt{HybridGen}}$ and $w_{\mathtt{VerifyGen}}$ achieve high performance ($73.53$ and $73.11$, respectively), applying them uniformly to all prompts incurs exorbitant costs ($\$6.27$ and $\$5.68$) and significant latency ($8.78$h and $6.94$h). By dynamically pruning unnecessary operations and selectively routing simple prompts to lightweight paths (\textit{e.g.}, $w_{\mathtt{RewriteGen}}$), \ourmethod achieves matching top-tier performance ($73.52$) while drastically slashing average cost to $\$1.37$ and latency to $1.76$h, positioning it in the optimal Pareto frontier.

\vspace{-0.4em}
\subsection{Evolution and Generalization Analysis}
\vspace{-0.4em}

\begin{wrapfigure}{r}{0.46\textwidth}
\vspace{-1.2em}
 \centering
 \includegraphics[width=\linewidth]{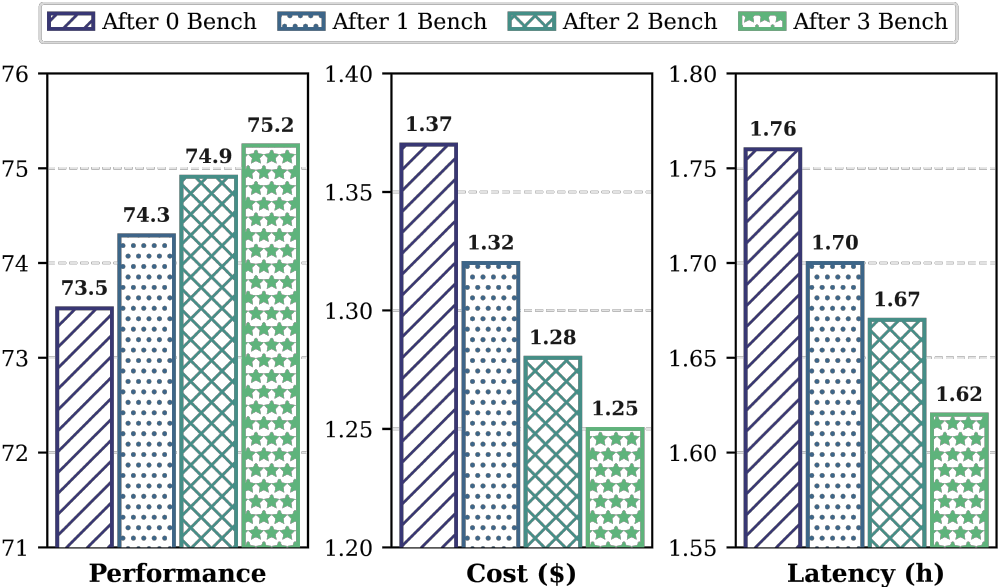}
  \vspace{-2em}
  \caption{Self-evolution analysis.}
  \vspace{-2em}
  \label{fig:figure4}
\end{wrapfigure}
To answer \textbf{RQ3}, we further assess the self-evolving capability and cross-benchmark generalization of \ourmethod. First, we evaluate its performance on the mixed test set after sequentially accumulating routing experience from up to three distinct benchmarks (OneIG-EN, OneIG-CN, and LongText-CN) in Figure~\ref{fig:figure4}. Furthermore, we perform a zero-shot domain transfer experiment on DPG-Bench, initializing the router solely with prior experience distilled from the WISE benchmark, as shown in Figure~\ref{fig:figure3} (\textit{Right}). Our findings are summarized below:

\textbf{Obs.\ding{186} \ourmethod effectively accumulates cross-task experience to continuously self-evolve, enabling zero-shot generalization across unseen distributions.}
As depicted in Figure~\ref{fig:figure4}, continuous exposure to new task distributions yields monotonic improvements across all metrics. After accumulating experience from three benchmarks, performance on the unseen mixed set improves from $73.5$ to $75.2$, while cost and latency drop by $8.7\%$ and $7.9\%$, respectively. Furthermore, Figure~\ref{fig:figure3} (\textit{Right}) demonstrates its superior generalization. When transferring from WISE to DPG-Bench, even with \textit{frozen} experience (disabling online updates), \ourmethod achieves an $87.1$ performance score, outperforming a standard LLM-as-Router ($86.4$) while halving the execution cost ($\$1.51$ \textit{vs.} $\$2.40$). Allowing \textit{updatable} experience during the target evaluation further boosts performance to $87.7$ and reduces cost to $\$1.34$. This confirms that our dual-memory architecture successfully distills task-agnostic routing priors that seamlessly generalize to new domains without catastrophic forgetting or the need for manual retraining.



\vspace{-1.4em}
\paragraph{Case Study \& Ablation Study.} Additional qualitative case studies demonstrating the framework's detailed execution steps are provided in Appendix \S\ref{app:a.4} (Figures~\ref{fig:app_fig_workflow_1}-\ref{fig:app_fig_case_2}). Furthermore, comprehensive ablation studies validating the necessity of the bounded workflow space, individual routing components, the dual-memory mechanism, and utility sensitivity are detailed in Appendix~\S\ref{app:more_results}.
\section{Conclusion}

In this work, we introduce \ourmethod, the first unified workflow routing framework for agentic image generation, alongside \ourspace, a standardized infrastructure of generative primitives and modular templates. By dynamically assigning heterogeneous prompts to optimal execution trajectories, our framework effectively mitigates the dual bottlenecks of system fragmentation and compute-mismatch inherent in existing ``one-size-fits-all" agentic pipelines. Through continuous experience distillation, \ourmethod autonomously self-evolves, enabling robust generalization across unseen domains. Extensive empirical validation demonstrates that \ourmethod within \ourspace not only achieves superior visual alignment but also drastically reduces execution cost and latency, establishing a highly efficient and scalable foundation for the next complex image generation.

\newpage
\bibliography{iclr_2027}
\bibliographystyle{plainnat}

\newpage
\appendix
\newpage

\appendix

\section{\ourspace Implementation}
\label{app:gencanvas_implementation}

\subsection{Formal Definition of Primitives}

\ourspace formalizes agentic workflows as directed compositions of primitive operators. For strict observability and cost tracking, each primitive emits a unified \texttt{PrimitiveTrace} logging the executed backend, token consumption, monetary cost, and latency. Furthermore, to cleanly decouple the execution logic from the underlying LLMs/MLLMs, \ourspace enforce strict JSON-based meta-prompt contracts for all agentic operations. The core foundational primitives are defined below:

\vspace{-0.6em}
\begin{itemize}[leftmargin=1.2em]
    \item \textbf{Decompose ($\pi_\mathsf{decompose}$):} This primitive translates the raw user prompt into a structured visual specification. It identifies explicit visible entities, verifiable constraints, and unresolved unknowns. The constraint parameters are subsequently converted into a checklist for the verification phase.
    \vspace{-0.6em}
\begin{tcolorbox}[notitle, sharp corners, breakable, colframe=CadetBlue!20, colback=gray!4, boxrule=2pt, boxsep=0.5pt, enhanced, shadow={3pt}{-3pt}{0pt}{opacity=1,gray!10}]\label{box:prompt}
       \footnotesize
       \setstretch{1}
       {\fontfamily{put}\selectfont
\begin{lstlisting}
{
  "entities": [{"id": "o1", "name": "visible entity", "priority": "primary"}],
  "constraints": [{
      "id": "c1", "text": "visual requirement", "type": "attribute"
  }],
  "unknowns": [{"id": "u1", "owner_id": "o1", "question": "what is unresolved?"}]
}
\end{lstlisting}
}
\end{tcolorbox}
\vspace{-0.6em}
    \item \textbf{Search ($\pi_\mathsf{search}$) \& Reason ($\pi_\mathsf{reason}$):} These primitives actively resolve the ``unknowns" identified during decomposition. $\pi_\mathsf{search}$ retrieves external textual facts or visual references, returning typed $\texttt{Evidence}$. $\pi_\mathsf{reason}$ resolves implicit visual implications that require logical, spatial, or causal deduction, returning its conclusions as reasoning notes.
    \item \textbf{Skill ($\pi_\mathsf{skill}$):} Functioning as a strategic router, this primitive evaluates the task signature against a built-in skill bank to retrieve reusable, domain-specific prompt-engineering instructions. If no specialized skill is matched, it gracefully falls back to lightweight heuristic guidelines for spatial and text rendering.
    \item \textbf{Sketch ($\pi_\mathsf{sketch}$):} For prompts demanding precise spatial layouts, this primitive generates an executable code sketch (\textit{e.g.}, SVG, HTML/CSS, or Three.js) to serve as a strict structural reference for the downstream image generator.
    \vspace{-0.6em}
\begin{tcolorbox}[notitle, sharp corners, breakable, colframe=CadetBlue!20, colback=gray!4, boxrule=2pt, boxsep=0.5pt, enhanced, shadow={3pt}{-3pt}{0pt}{opacity=1,gray!10}]\label{box:prompt}
       \footnotesize
       \setstretch{1}
       {\fontfamily{put}\selectfont
\begin{lstlisting}
{
  "reasoning": ["brief planning note"],
  "sketch_type": "svg | html_css | threejs",
  "records": [{"id": "r1", "kind": "entity", "details": "what and where"}],
  "code": "complete directly renderable code",
  "render_prompt": "Follow this sketch as a strict layout reference."
}
\end{lstlisting}
}
\end{tcolorbox}
\vspace{-0.6em}
    \item \textbf{Verify ($\pi_\mathsf{verify}$):} Powered by an MLLM backend, this primitive closes the execution loop by evaluating the generated image against the checklist derived from $\pi_\mathsf{decompose}$. It assigns failure families (\textit{e.g.}, layout, attribute, count) to unmet constraints to guide the subsequent refinement.
    \vspace{-0.6em}
\begin{tcolorbox}[notitle, sharp corners, breakable, colframe=CadetBlue!20, colback=gray!4, boxrule=2pt, boxsep=0.5pt, enhanced, shadow={3pt}{-3pt}{0pt}{opacity=1,gray!10}]\label{box:prompt}
       \footnotesize
       \setstretch{1}
       {\fontfamily{put}\selectfont
\begin{lstlisting}
{
  "items": [{
      "constraint_id": "c1", 
      "passed": false, 
      "rationale": "short visual reason",
      "failure_family": "layout"
  }]
}
\end{lstlisting}
}
\end{tcolorbox}
\vspace{-0.6em}
    \item \textbf{Refine ($\pi_\mathsf{refine}$) \& Rewrite ($\pi_\mathsf{rewrite}$):} $\pi_\mathsf{rewrite}$ synthesizes the original prompt, retrieved evidence, skill instructions, and reasoning notes into a self-contained, generator-ready prompt. In the event of a verification failure, $\pi_\mathsf{refine}$ ingests the $\texttt{VerificationFeedback}$ and attempt history to formulate a corrected prompt, explicitly instructed to fix failed targets while preserving passed requirements.
\end{itemize}
\vspace{-0.6em}

By enforcing these standardized input-output interfaces, \ourspace ensures that complex cognitive tasks are reliably decomposed into manageable, trackable, and heavily constrained atomic operations.

\subsection{Topological Configurations of Workflow}

In \ourspace, workflow templates are implemented as modular Python classes whose execution methods compose primitives into directed topologies. This design cleanly decouples the execution logic from the underlying generator backends. Below, we detail the structural configurations of two representative workflows:
\vspace{-0.6em}
\begin{itemize}[leftmargin=1.2em]
\item \textbf{Iterative Refinement} ($w_\mathtt{VerifyGen}$):
This template implements a closed-loop topology to systematically correct generation failures. It begins by invoking $\pi_\mathsf{decompose}$ to extract the verification checklist and $\pi_\mathsf{rewrite}$ to formulate the initial prompt. The workflow then enters a loop bounded by a predefined $\texttt{max\_iter}$. In each iteration, it executes the generator $g$ and immediately evaluates the output using $\pi_\mathsf{verify}$. If constraints fail, the attempt history and feedback are passed to $\pi_\mathsf{refine}$ to update the prompt for the subsequent iteration.
\item \textbf{Structural Reasoning ($w_\mathtt{SketchGen}$)}:
To handle demanding spatial constraints, this topology separates symbolic layout construction from photorealistic synthesis. It first utilizes $\pi_\mathsf{sketch}$ to generate deterministic, executable visual code (\textit{e.g.}, SVG or HTML/CSS) representing the spatial layout. The image generator then ingests the rendered sketch as a strict structural reference, focusing solely on adding style, texture, lighting, and realism to the explicitly defined layout.
\end{itemize}
\vspace{-0.6em}

\subsection{Extensibility and Architecture}

To facilitate rapid prototyping and rigorous benchmarking, \ourspace is architected with strict modularity, allowing seamless integration of new primitives, topologies, and backends.

\vspace{-1.4em}
\paragraph{Decoupling Logic from Backend Models.}
A critical design principle of \ourspace is the strict separation of logical topologies from proprietary model APIs. All primitives interact exclusively with standardized abstract protocols (\textit{e.g.}, \texttt{LLMBackend}, \texttt{GeneratorBackend}). This agnostic design allows researchers to seamlessly swap providers, from local pipelines to commercial endpoints, without altering core workflow logic. Concrete models are managed via a Registry pattern and configured through a YAML interface, which automatically enforces capability constraints.

\vspace{-1.4em}
\paragraph{Unified Evaluation and Accounting Engine.}
Evaluating agentic systems requires tracking both visual fidelity and computational overhead. GenCanvas features an integrated evaluation scaffold that standardizes accounting across all workflows. Every primitive execution logs its specific token usage, API pricing, and execution latency into a \texttt{PrimitiveTrace}. The runner then aggregates these traces into a unified output:
\vspace{-0.6em}
\begin{tcolorbox}[notitle, sharp corners, breakable, colframe=CadetBlue!20, colback=gray!4, boxrule=2pt, boxsep=0.5pt, enhanced, shadow={3pt}{-3pt}{0pt}{opacity=1,gray!10}]\label{box:prompt}
       \footnotesize
       \setstretch{1}
       {\fontfamily{put}\selectfont
\begin{lstlisting}
WorkflowResult(
    workflow=str,          # e.g., "HybridGen"
    generator=str,         # e.g., "Qwen-Image-2512"
    trace=list[PrimitiveTrace],
    score=float,           # Quality metric
    cost=float,            # Aggregated monetary cost
    latency=float,         # End-to-end execution time
)
\end{lstlisting}
}
\end{tcolorbox}
\vspace{-0.2em}

\begin{figure*}[!b]
\centering
\vspace{-0.4em}
\includegraphics[width=\linewidth]{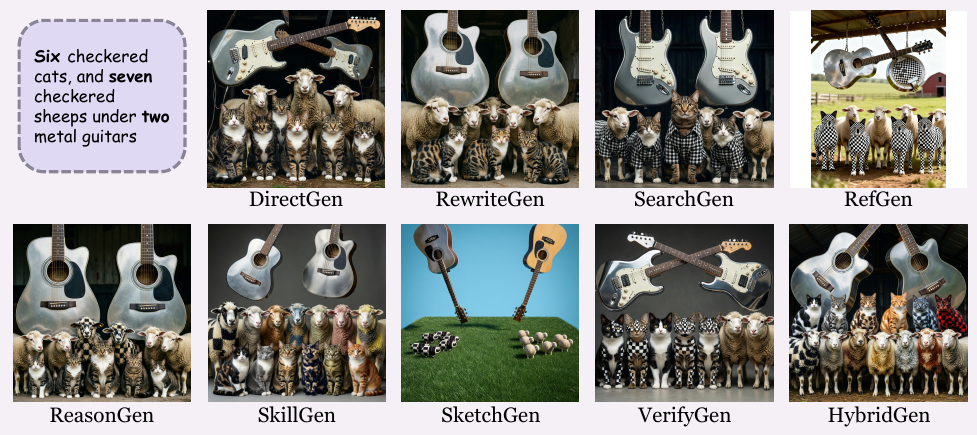}
\vspace{-1.8em}
\caption{Workflow demonstration of \ourspace \& \ourmethod.}
\label{fig:app_fig_workflow_1}
\vspace{-0.6em}
\end{figure*}

\begin{figure*}[!b]
\centering
\includegraphics[width=\linewidth]{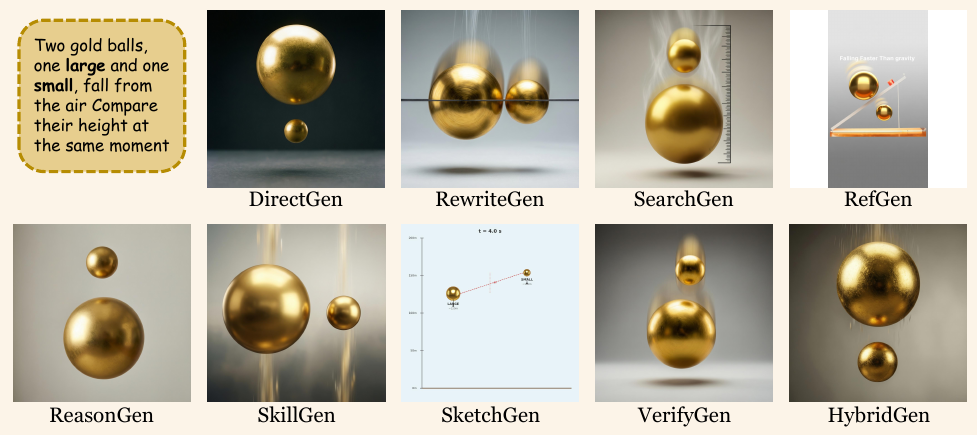}
\vspace{-1.8em}
\caption{Workflow demonstration of \ourspace \& \ourmethod.}
\label{fig:app_fig_workflow_2}
\vspace{-0.8em}
\end{figure*}

\subsection{Showcase}
\label{app:a.4}

In this section, we provide detailed visual examples to further illustrate the generative capabilities and routing mechanisms of \ourmethod within \ourspace.

\vspace{-1.4em}
\paragraph{Workflow Demonstrations.} Figures~\ref{fig:app_fig_workflow_1} and \ref{fig:app_fig_workflow_2} showcase the visual outcomes of a single prompt executed across the entire spectrum of \ourspace workflows. These examples clearly demonstrate how varying degrees of cognitive intervention, from naive direct generation to iterative hybrid refinement, drastically alter the structural composition, factual accuracy, and semantic alignment of the final image.

\begin{figure*}[!h]
\centering
\vspace{-0.4em}
\includegraphics[width=\linewidth]{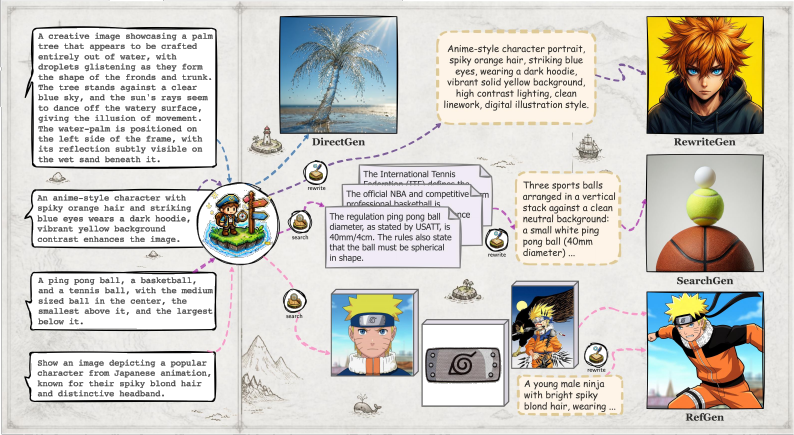}
\vspace{-1.8em}
\caption{Case study of \ourspace \& \ourmethod.}
\label{fig:app_fig_case_1}
\vspace{-1em}
\end{figure*}

\begin{figure*}[!h]
\centering
\vspace{-0.4em}
\includegraphics[width=\linewidth]{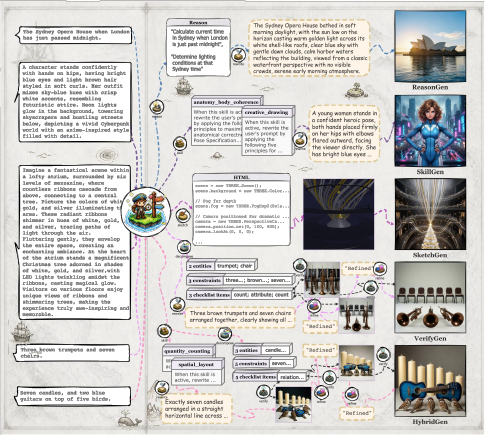}
\vspace{-1.8em}
\caption{Case study of \ourspace \& \ourmethod.}
\label{fig:app_fig_case_2}
\vspace{-1em}
\end{figure*}

\vspace{-1.4em}
\paragraph{Case Studies.} Figures \ref{fig:app_fig_case_1} and \ref{fig:app_fig_case_2} provide a granular look at the internal execution logic of \ourmethod. For representative prompts, we visualize the step-by-step trajectory, detailing how \ourmethod intelligently dispatches specific primitives (\textit{e.g.}, extracting factual knowledge via $\pi_{\mathsf{search}}$, rendering structural code via $\pi_{\mathsf{sketch}}$, or enforcing constraints via iterative $\pi_{\mathsf{verify}}$ loops) to systematically resolve complex generative intents.

\section{\ourmethod Implementation}
\label{app:genrouter_implementation}

\subsection{Cold Start and Experience Distillation}

To bridge the gap between deterministic priors and environment-specific performance, \ourmethod employs a cold-start initialization phase (isolated from the evaluation process) to construct empirical baselines for its dual-memory system.

\vspace{-1.4em}
\paragraph{Exhaustive Exploration.} We initialize the cold-start phase with $N=10$ diverse calibration prompts sampled from the target benchmark. For each prompt, the demand profiler extracts its task signature to construct compatible workflow-generator candidate plans. \ourmethod exhaustively executes all valid candidates for these $N$ prompts, ensuring unbiased data collection across all generative trajectories.

\vspace{-1.4em}
\paragraph{Batched Evaluation.} Rather than scoring generations on-the-fly, \ourmethod uniformly evaluates accumulated outputs every $50$ executions to optimize computational overhead. We employ official metrics and scorer backends from the targeted benchmarks (\textit{e.g.}, WISE \citep{niu2025wise}, DPG-Bench \citep{hu2024ella}) to assess visual performance. These scores are then synthesized with automatically logged token consumption $C$ and execution time $L$ to compute the final scalarized utility.

\vspace{-1.4em}
\paragraph{Periodic Memory Distillation.} Following each batched evaluation (every $50$ records), the evaluated prompt-plan outcomes are appended to the trajectory memory $\mathcal{M}_\text{traj}$. Concurrently, \ourmethod distills these records into the route memory $\mathcal{M}_\text{route}$ by grouping them into coarse task buckets based on their dominant signature requirements. Within each bucket, aggregate statistics (\textit{e.g.}, mean score, cost, and latency) are updated, and Pareto-optimal plans are dynamically recalculated. This periodic evaluation-distillation loop operates continuously during both the cold-start phase and online deployment, empowering \ourmethod to autonomously self-evolve.

\subsection{More Details of Demand Profiling}

As introduced in Section~\S\ref{sec:5.1}, \ourmethod employs a lightweight LLM (\textit{i.e.}, Qwen3.5-4B \citep{yang2025qwen3}) as a demand profiler to map the incoming user prompt into a $7$-dimensional task signature $z(x)$. The detailed system prompt template is provided below:
\vspace{-0.6em}
\begin{tcolorbox}[notitle, sharp corners, breakable, colframe=teal!40, colback=gray!4, 
       boxrule=3pt, boxsep=0.5pt, enhanced, 
       shadow={3pt}{-3pt}{0pt}{opacity=1,gray!10},
       title={Prompt for Signature Extraction}]\label{box:prompt}
       \footnotesize
       \setstretch{1}
       {\fontfamily{pcr}\selectfont
\begin{lstlisting}
"""
Analyze the user's image generation prompt, briefly explain the scoring rationale, and then output a task signature as JSON.

Output format:
<reason>
Semantic_Articulation: [score], [a short explanation] 
Factual_Grounding: [score], [a short explanation]
Visual_Referencing: [score], [a short explanation]
Logical_Deduction: [score], [a short explanation]
Compositional_Heuristics: [score], [a short explanation]
Evaluative_Critique: [score], [a short explanation]
Spatial_Layout: [score], [a short explanation]

Give the score and then briefly explain from each field. Keep the explanation concise and grounded in the given prompt. 
</reason>

<json>
{"semantic_articulation": "...",
 "factual_grounding": "...",
 "visual_referencing": "...",
 "logical_deduction": "...",
 "compositional_heuristics": "...",
 "evaluative_critique": "...",
 "spatial_layout": "..."} 
Each value must be a numeric score in 0, 1, 2, 3, 4, or 5. 
</json>

"""
\end{lstlisting}
}
\end{tcolorbox}
\vspace{-0.4em}

\begin{table*}[!b]
\centering
\vspace{-0.6em}
  \caption{Composition of the mixed test set.}
    \vspace{-0.8em}
    \renewcommand{\arraystretch}{1.3}
    \resizebox{\textwidth}{!}{%
   \begin{tabular}{l|ccccccccc}
     \hlineB{2.5}
    \rowcolor{CadetBlue!20} 
    \textbf{Info.} & \textbf{WISE} & \textbf{DPG} & \textbf{OneIG-EN} & \textbf{OneIG-CN} & \textbf{LongText-EN} & \textbf{LongText-CN} & \textbf{SpatialGenEval} & \textbf{ArtiMuse} & \textbf{GenEval2}\\
    \hlineB{1.5}
    \textbf{Quantity} & 100 & 50 & 50 & 50 & 50 & 50 & 50 & 50 & 50 \\
    \rowcolor{gray!10}
    \textbf{Score Weight} & 1.12 & 1.21 & 1.44 & 1.43 & 1.03 & 1.05 & 1.63 & 1.69 & 1.59  \\
     \hlineB{2.5}
   \end{tabular}%
}
  \label{app_tab:mixed_set}
  \vspace{-0.6em}
\end{table*}

\subsection{Algorithm Workflow}

We summarize the overall dynamic routing workflow of \ourmethod in Algorithm~\ref{algo:genrouter}.

\newcommand\mycommfont[1]{\textcolor{teal}{\textit{#1}}}

\begin{algorithm}[!h]\small
\DontPrintSemicolon
\SetAlgoLined
\LinesNumbered
\SetCommentSty{mycommfont} 

\caption{\ourmethod}
\label{algo:genrouter}
\KwIn{Prompt $x$, Workflows $\mathcal{W}$, Generators $\mathcal{G}$, Memories $\mathcal{M}_{\text{traj}}, \mathcal{M}_{\text{route}}$, Penalties $\lambda_c, \lambda_l$}
\KwOut{Generated Image $y$}

\tcp{Stage I: Demand Profiling \& Candidate Pruning}
$z(x) \gets \text{LLM-Profiler}(x)$ \tcp*{Extract 7-D task signature}
$\mathcal{P}_{\text{valid}} \gets \emptyset$ \;
\ForEach{$w \in \mathcal{W}$ \textbf{and} $g \in \mathcal{G}$}{
  \If{$(w, g)$ satisfies capability and gating constraints given $z(x)$}{
    $\mathcal{P}_{\text{valid}} \gets \mathcal{P}_{\text{valid}} \cup \{p=(w, g)\}$ \;
  }
}

\tcp{Stage II: Memory-guided Matching}
\ForEach{$p \in \mathcal{P}_{\text{valid}}$}{
  $(\hat{S}_{\text{traj}}, \hat{C}_{\text{traj}}, \hat{L}_{\text{traj}}) \gets \text{Retrieve}(\mathcal{M}_{\text{traj}}, p, z(x), k)$ \;
  $(\hat{S}_{\text{route}}, \hat{C}_{\text{route}}, \hat{L}_{\text{route}}) \gets \text{Aggregate}(\mathcal{M}_{\text{route}}, p)$ \;
  
  \eIf{$\exists$ historical memory for $p$}{
    Compute confidence $\alpha_p \in [0, \alpha_0]$ based on match count \;
    $\hat{S}_p \gets \alpha_p \hat{S}_{\text{traj}} + (1 - \alpha_p) \hat{S}_{\text{route}}$ \tcp*{Similarly for $\hat{C}_p, \hat{L}_p$}
  }{
    $(\hat{S}_p, \hat{C}_p, \hat{L}_p) \gets \text{Prior}(z(x), p)$ \tcp*{Deterministic fallback}
  }
  $\hat{U}_p \gets \hat{S}_p - \lambda_c \hat{C}_p - \lambda_l \hat{L}_p$ \;
}

\tcp{Stage III: Pareto Filtering \& Execution}
$\mathcal{P}_{\text{pareto}} \gets \emptyset$ \;
\ForEach{$p \in \mathcal{P}_{\text{valid}}$}{
  \If{$\nexists \; p' \in \mathcal{P}_{\text{valid}} \setminus \{p\}$ dominating $p$ across $(\hat{S}, -\hat{C}, -\hat{L})$}{
    $\mathcal{P}_{\text{pareto}} \gets \mathcal{P}_{\text{pareto}} \cup \{p\}$ \;
  }
}
$p^* \gets \arg\max_{p \in \mathcal{P}_{\text{pareto}}} \hat{U}_p$ \;
$y \gets p^*(x)$ \;
\Return{$y$}
\end{algorithm}

\section{Experimental Details}
\label{app:experimental_details}

\subsection{Evaluation Details}
\label{app:c.1}

\paragraph{Metrics.} We evaluate the framework based on three dimensions: performance, cost, and latency. Performance corresponds to the official scoring mechanism of each respective benchmark. For cost and latency, we exclusively measure the overhead incurred by the routing process and primitive execution (\textit{e.g.}, LLM reasoning or search APIs). We deliberately \textbf{exclude} the computational cost and inference latency of the image generator. Isolating these factors is crucial for unbiased generator routing; otherwise, the inherent disparities in generator inference times and costs could heavily skew the utility function and cause an unfair bias against more capable, heavyweight models. The strategy for standardizing these performance metrics across heterogeneous tasks is detailed below.

\vspace{-1.4em}
\paragraph{The Mixed Test Set.} To evaluate the robustness and generalization capabilities of \ourmethod across highly diverse generative scenarios, we construct a comprehensive mixed test set. As detailed in Table \ref{app_tab:mixed_set}, this set aggregates prompts from nine distinct benchmark distributions to ensure broad coverage of primitive demands. However, aggregating heterogeneous benchmarks introduces a significant challenge: each dataset employs distinct scoring scales and emphasizes completely different generative capabilities (\textit{e.g.}, spatial accuracy \textit{vs.} text rendering). To address this, we design specific score weights to calibrate each subset. This normalization aligns the disparate evaluation metrics into a unified reward signal, enabling seamless experience sharing and allowing the router to effectively accumulate cross-task routing priors.

\subsection{Baseline Details}

In this section, we provide implementation configurations for each baseline method included in our comparison. To ensure a fair and rigorous evaluation, we standardize the underlying LLM/MLLM backbone across all baseline workflows to Kimi K2.5 \citep{team2026kimi} (the default in GEMS \citep{he2026gems}), eliminating performance variances caused by the proprietary models originally used. 

\vspace{-0.6em}
\begin{itemize}[leftmargin=1.2em]
\item \textbf{GEMS} \citep{he2026gems} orchestrates an iterative refinement loop with persistent memory. It translates the prompt into atomic criteria ($\pi_\mathsf{decompose}$) and enters a closed-loop process where the generated image ($g$) is evaluated by a multimodal verifier ($\pi_\mathsf{verify}$). Evaluation failures guide a refiner agent ($\pi_\mathsf{refine}$) to update the prompt iteratively. Within \ourspace, this maps to the $w_\mathtt{VerifyGen}$ and $w_\mathtt{SkillGen}$ templates. 
\item \textbf{SCOPE} \citep{ren2026scope} utilizes a specification-guided skill orchestration framework to maintain semantic commitments across the generation lifecycle. It translates requests into a structured semantic specification and conditionally invokes retrieval ($\pi_\mathsf{search}$), reasoning ($\pi_\mathsf{reason}$), and repair ($\pi_\mathsf{refine}$) skills to resolve unknowns and fix localized failures identified by an itemized verifier ($\pi_\mathsf{verify}$). This aligns with our comprehensive $w_\mathtt{HybridGen}$ template. 
\item \textbf{Mind-Brush} \citep{he2026mind} employs an agentic ``think-research-create" workflow. It detects cognitive gaps and actively retrieves multimodal external evidence ($\pi_\mathsf{search}$) while leveraging chain-of-thought logic ($\pi_\mathsf{reason}$) to resolve implicit constraints. A review agent then consolidates the evidence into an enriched master prompt ($\pi_\mathsf{rewrite}$) prior to synthesis ($g$). This maps directly to our external grounding templates, $w_\mathtt{SearchGen}$ and $w_\mathtt{RefGen}$. 
\end{itemize}
\vspace{-0.6em}

\section{More Results \& Sensitivity Analysis}
\label{app:more_results}

\subsection{Workflow \textit{vs.} Primitive Routing}
\begin{wrapfigure}{r}{0.34\textwidth}
\vspace{-4em}
 \centering
 \includegraphics[width=\linewidth]{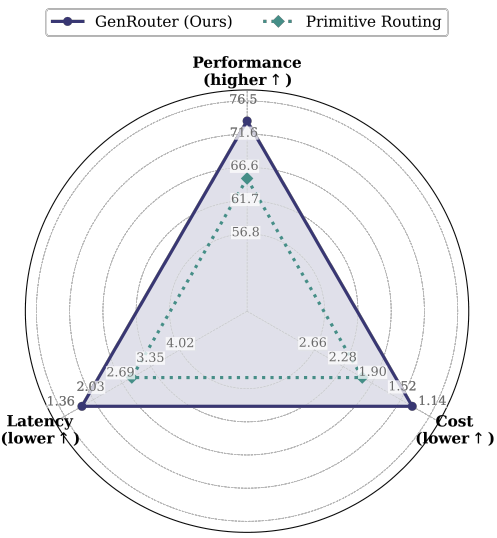}
  \vspace{-2em}
  \caption{Analysis of bounded space of \ourspace.}
  \vspace{-3em}
  \label{app_fig:e.1}
\end{wrapfigure}

To validate the necessity of \ourspace's bounded templates, we compare \ourmethod against ``Primitive Routing", where the router is given unconstrained access to the raw primitive library $\Pi$ without predefined topologies. As illustrated in Figure~\ref{app_fig:e.1}, unconstrained routing leads to significant execution instability and redundant tool invocations. It achieves a lower performance of $65.00$ while incurring substantially higher costs ($\$2.03$) and latency ($2.91$h) compared to \ourmethod's $73.52$ performance at $\$1.37$ and $1.76$h. This confirms that \ourspace provides a critical stabilizing infrastructure, preventing the router from getting trapped in inefficient or infinite loops.

\subsection{Component Ablation of \ourmethod}

\begin{wrapfigure}{r}{0.34\textwidth}
\vspace{-2.6em}
 \centering
 \includegraphics[width=\linewidth]{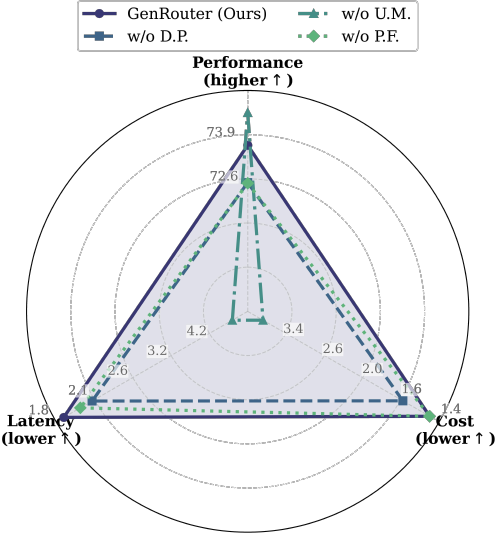}
  \vspace{-2em}
  \caption{Analysis of components of \ourmethod.}
  \vspace{-4.4em}
  \label{app_fig:e.2}
\end{wrapfigure}

We further ablate the core components of \ourmethod in Figure~\ref{app_fig:e.2}. Removing Demand Profiling (w/o D.P.) and routing directly based on prompt embeddings degrades performance to $72.54$ and increases overhead, highlighting the importance of explicitly extracting cognitive signatures. Disabling Utility Matching (w/o U.M.) by clearing the empirical memories drastically spikes costs to $\$5.33$ and latency to $7.13$h, as the router falls back to naive, uncalibrated thresholding. Finally, removing Pareto Filtering (w/o P.F.) leads to sub-optimal plan selections, slightly lowering performance ($72.52$) and increasing latency ($1.93$h). The integration of all three components is essential for achieving the optimal efficiency-performance balance.

\subsection{Utility Analysis}

\begin{wraptable}{r}{0.43\textwidth}
 \vspace{-1.34em}
 \centering
  \centering
  \caption{Sensitivity analysis of \ourmethod +\qwenlogo~~on the mixed test set.}
  \label{app_tab:sensitivity}
  \vspace{-0.8em}
  \renewcommand\tabcolsep{2.6pt}
  \renewcommand\arraystretch{1.2}
  \scriptsize
  \begin{tabular}{lccc} 
    \hlineB{2.5}
    \rowcolor{CadetBlue!20} 
    \textbf{Utility}  & \textbf{Performance$\uparrow$} & \textbf{Cost (\$)$\downarrow$} & \textbf{Latency (h)$\downarrow$}\\
    \hlineB{1.5}
    $\lambda_c=1.0$ & 73.56 & 1.56 & 1.98 \\
    \rowcolor{gray!10}
    $\lambda_c=5.0$ & 73.52 & 1.37 & 1.76  \\
    $\lambda_c=10.0$ & 73.00 & 1.36 & 1.87  \\
    \hline
    \rowcolor{gray!10}
    $\lambda_l=0.0003$ & 73.26 & 1.50 & 1.85 \\
    $\lambda_l=0.0006$ & 73.52 & 1.37 & 1.76  \\
    \rowcolor{gray!10}
    $\lambda_l=0.0010$ & 73.47 & 1.37 & 1.74  \\
    \hlineB{2}
  \end{tabular}
  \vspace{-2em}
\end{wraptable}

In this section, we investigate the sensitivity of \ourmethod to the utility trade-off coefficients, specifically the cost penalty $\lambda_c$ and the latency penalty $\lambda_l$, as defined in Eq. (\ref{eq:utility}). The results on the mixed test set are summarized in Table~\ref{app_tab:sensitivity}.

\vspace{-1.4em}
\paragraph{Cost Coefficient $\lambda_c$.} We evaluate \ourmethod by varying $\lambda_c \in \{1.0, 5.0, 10.0\}$ while keeping $\lambda_l$ fixed at $0.0006$. As shown in Table~\ref{app_tab:sensitivity} (\textit{Top}), increasing $\lambda_c$ from $1.0$ to $5.0$ significantly reduces both the execution cost (from $1.56$ to $1.37$) and latency, with only a negligible impact on visual performance ($73.56$ to $73.52$). However, applying an overly aggressive cost penalty ($\lambda_c=10.0$) forces the router into overly simplistic workflows, leading to a noticeable performance degradation ($73.00$) with diminishing returns in cost savings. 
Therefore, we adopt $\lambda_c=5.0$ as the optimal default configuration to maintain high generation quality while strictly bounding API expenses.

\vspace{-1.4em}
\paragraph{Latency Coefficient $\lambda_l$.} We vary $\lambda_l \in \{0.0003, 0.0006, 0.0010\}$ while keeping $\lambda_c$ fixed at $5.0$. An insufficient latency penalty ($\lambda_l=0.0003$) results in a sub-optimal routing strategy that tolerates redundant primitive invocations, yielding both lower performance ($73.26$) and higher execution overhead. Increasing the penalty to $0.0006$ achieves the best Pareto-optimal balance across all three metrics. Further increasing $\lambda_l$ to $0.0010$ marginally reduces latency ($1.76$ to $1.74$ hours) but slightly compromises generation quality. 
Hence, $\lambda_l=0.0006$ is selected to effectively penalize prohibitive execution delays without sacrificing the visual integrity of complex requests.

\subsection{Dual-Memory Experience Analysis}

\begin{wraptable}{r}{0.426\textwidth}
 \vspace{-1.34em}
 \centering
  \centering
  \caption{Memory analysis of \ourmethod +\qwenlogo~~on the mixed test set.}
  \label{app_tab:memory}
  \vspace{-0.8em}
  \renewcommand\tabcolsep{2.6pt}
  \renewcommand\arraystretch{1.2}
  \scriptsize
  \begin{tabular}{lccc} 
    \hlineB{2.5}
    \rowcolor{CadetBlue!20} 
    \textbf{Method}  & \textbf{Performance$\uparrow$} & \textbf{Cost (\$)$\downarrow$} & \textbf{Latency (h)$\downarrow$}\\
    \hlineB{1.5}
    Original \qwenlogo & 62.51 & 0 & 0 \\
    \hline
    \rowcolor{gray!10}
    + $\mathcal{M}_\text{traj}$ only & 72.18 & 1.29 & 1.71  \\
    + $\mathcal{M}_\text{route}$ only & 72.15 & 1.13 & 1.57  \\
    \rowcolor{gray!10}
    + \ourmethod & 73.52 & 1.37 & 1.76  \\
    \hlineB{2}
  \end{tabular}
  \vspace{-1.4em}
\end{wraptable}

To validate the efficacy of our proposed dual-memory matching mechanism, we ablate the trajectory memory ($\mathcal{M}_{\text{traj}}$) and route memory ($\mathcal{M}_{\text{route}}$) on the mixed test set. The quantitative results are detailed in Table~\ref{app_tab:memory}.

Compared to the original static generator, which yields a baseline performance of $62.51$ with zero additional framework cost and latency, introducing either memory module independently yields substantial improvements. Relying solely on instance-level retrieval ($+\mathcal{M}_{\text{traj}}$ only) boosts performance to $72.18$, incurring a cost of $1.29$ and a latency of $1.71$h. Meanwhile, utilizing only the distilled bucket-level statistics ($+\mathcal{M}_{\text{route}}$ only) achieves a comparable performance score of $72.15$, but with noticeably lower execution overhead at a cost of $1.13$ and latency of $1.57$h. This highlights the effectiveness of $\mathcal{M}_{\text{route}}$ in smoothing out noisy execution traces and providing a robust, computationally efficient prior.
Crucially, the full \ourmethod framework, which dynamically calibrates both memories, achieves the highest overall performance of $73.52$. While the combined system incurs marginally higher computational overhead ($1.37$ cost, $1.76$h latency) compared to the single-memory variants, the significant performance gain justifies the trade-off. This underscores the synergistic relationship between the two components: $\mathcal{M}_{\text{traj}}$ offers precise, fine-grained routing for familiar queries, whereas $\mathcal{M}_{\text{route}}$ mitigates data sparsity and acts as a stable anchor, collectively guiding the router toward the optimal execution plan.

\section{Exhibition Board}
\label{app:exhibition}

\begin{figure*}[!h]
\centering
\includegraphics[width=1\linewidth]{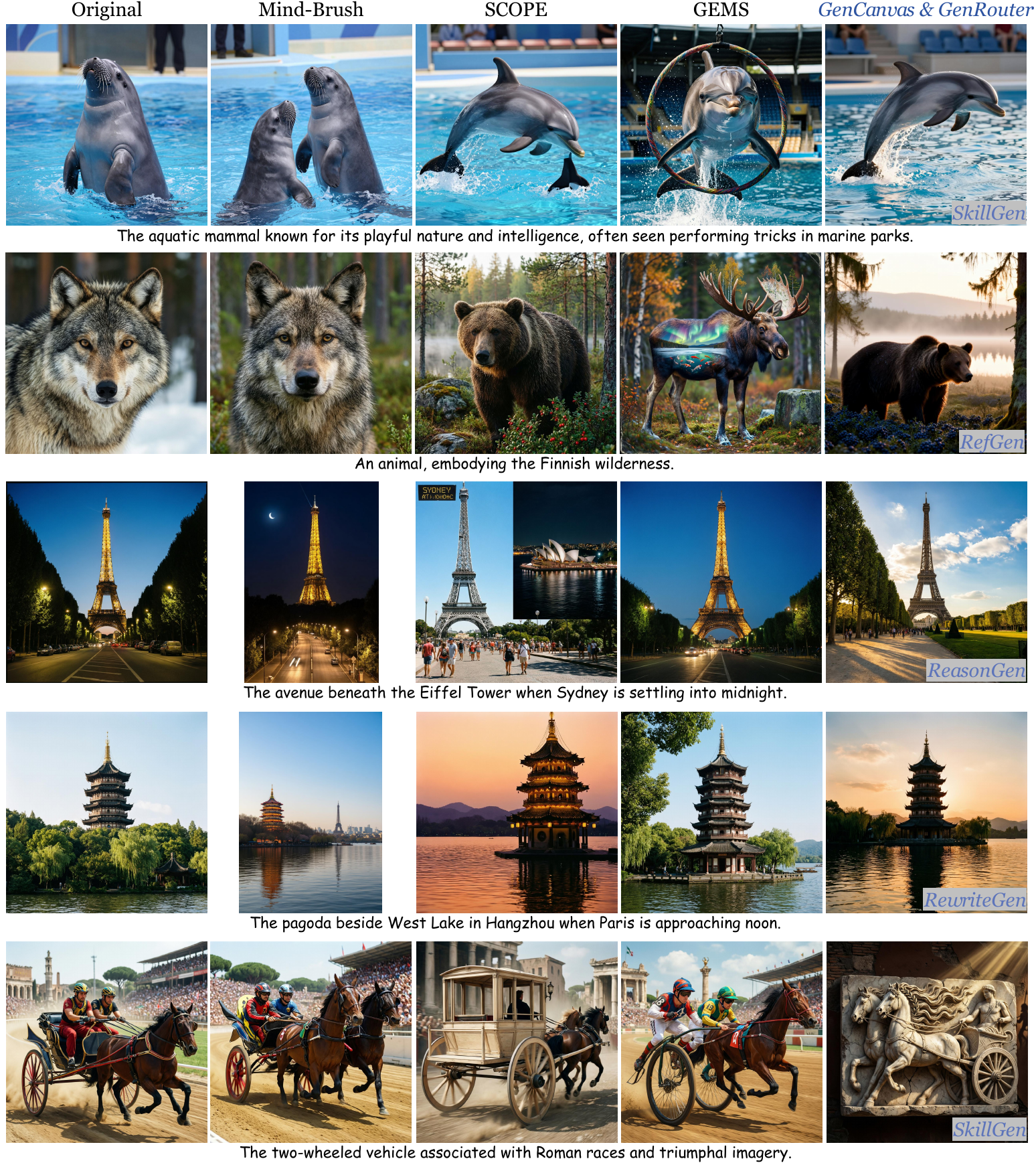}
\vspace{-1.4em}
\caption{More results demonstration of \ourmethod within \ourspace.
}
\label{fig:exhibition}
\end{figure*}

\begin{figure*}[!h]
\centering
\includegraphics[width=1\linewidth]{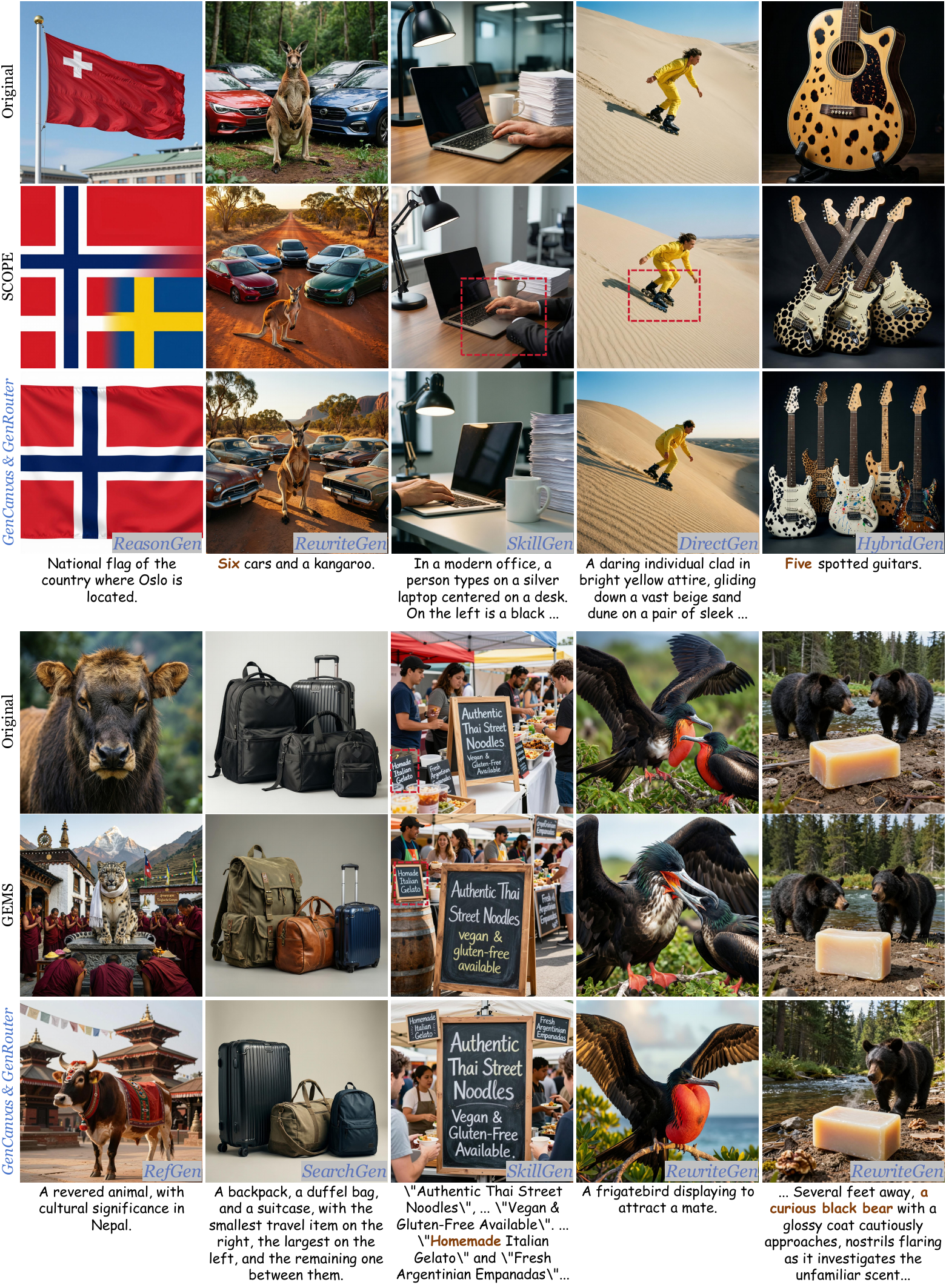}
\vspace{-1.4em}
\caption{More results demonstration of \ourmethod within \ourspace.
}
\label{fig:exhibition_2}
\end{figure*}

We provide more comparison results here in Figures~\ref{fig:exhibition} and \ref{fig:exhibition_2}.



\end{document}